\documentclass[1p,sort&compress,numbers]{elsarticle}

\usepackage{graphicx}
\usepackage{amssymb}
\usepackage{amsmath}
\usepackage[utf8]{inputenc}
\usepackage{pifont}
\usepackage[nolist]{acronym}
\usepackage{tabularx}
\usepackage[noabbrev,sort&compress]{cleveref}
\usepackage{url}
\usepackage{subcaption}
\usepackage[dvipsnames]{xcolor}
\usepackage{algorithm2e}
\usepackage{fontawesome}
\usepackage{subcaption}
\usepackage{tikz}

\usepackage{float}

\usepackage{enumitem}

\journal{Information Sciences}

\newcommand{\cL}{\mathcal{L}}
\newcommand{\cF}{\mathcal{F}}
\newcommand{\cl}{\mathcal{\ell}}
\newcommand{\cmark}{\ding{51}}
\newcommand{\xmark}{\ding{55}}

\begin{acronym}
    \acro{RS}{Recommender System}
    \acro{CF}{Collaborative Filtering}
    \acro{MF}{Matrix Factorization}
    \acro{MAE}{Mean Absolute Error}
    \acro{MSE}{Mean Squared Error}
    \acro{RMSE}{Root Mean Squared Error}
    \acro{CF4J}{Collaborative Filtering for Java}
    \acro{BeMF}{Bernoulli Matrix Factorization}
    \acro{RPI}{Reliability Prediction Index}
    \acro{KNN}{K Nearest Neighbours}
    \acro{PMF}{Probabilistic Matrix Factorization}
    \acro{NMF}{non-Negative Matrix Factorization}
    \acro{BNMF}{Bayesian non-Negative Matrix Factorization}
    \acro{URP}{User Ratings Profile Model}
    \acro{NN}{Neural Network}
\end{acronym}

\begin{document}

\newtheorem{thm}{Theorem}[section]
\newtheorem{prop}[thm]{Proposition}
\newtheorem{lem}[thm]{Lemma}
\newtheorem{cor}[thm]{Corollary}

\newtheorem{defn}[thm]{Definition}
\newtheorem{as}{Assumption}

\newtheorem{rmk}[thm]{Remark}
\newtheorem{ex}[thm]{Example}

\usetikzlibrary{bayesnet}
\usetikzlibrary{arrows}

\begin{frontmatter}
    \title{Providing Reliability in Recommender Systems through Bernoulli Matrix Factorization}
    \author[1]{Fernando Ortega\corref{cor1}}\ead{fernando.ortega@upm.es}
    \author[1]{Raúl Lara-Cabrera}\ead{raul.lara@upm.es}
    \author[1]{Ángel González-Prieto}\ead{angel.gonzalez.prieto@upm.es}
    \author[1]{Jesús Bobadilla}\ead{jesus.bobadilla@upm.es}

    \cortext[cor1]{Corresponding author}
    \address[1]{Dpto. Sistemas Informáticos, ETSI Sistemas Informáticos, Universidad Politécnica de Madrid, Madrid, Spain}
    
    \begin{abstract}
        Beyond accuracy, quality measures are gaining importance in modern recommender systems, with reliability being one of the most important indicators in the context of collaborative filtering. This paper proposes Bernoulli Matrix Factorization (BeMF), which is a matrix factorization model, to provide both prediction values and reliability values. BeMF is a very innovative approach from several perspectives: a) it acts on model-based collaborative filtering rather than on memory-based filtering, b) it does not use external methods or extended architectures, such as existing solutions, to provide reliability, c) it is based on a classification-based model instead of traditional regression-based models, and d) matrix factorization formalism is supported by the Bernoulli distribution to exploit the binary nature of the designed classification model. The experimental results show that the more reliable a prediction is, the less liable it is to be wrong: recommendation quality improves after the most reliable predictions are selected. State-of-the-art quality measures for reliability have been tested, which shows that BeMF outperforms previous baseline methods and models.
    \end{abstract}
    
\end{frontmatter}

\section{Introduction}\label{sec:introduction}

\acp{RS}~\cite{Saquib2017} are services that most people use. Relevant commercial examples are Netflix, Spotify, Amazon and TripAdvisor. Improving recommendation accuracy has focused existing research~\cite{Bobadilla2013Jul}, and prediction models, such as the \ac{KNN} algorithm~\cite{Bobadilla2013Jul} to the \ac{MF} methods~\cite{Li2019Nov} and the latest \ac{NN} models~\cite{Batmaz2019Jun, Bobadilla2020Jan} have been refined over time. In recent years, increasing importance has been given to ``beyond accuracy'' \ac{RS} measures~\cite{Kaminskas2016Dec}. Diversity, coverage and serendipity are current goals in \acp{RS}, where diverse and innovative recommendations are highly desirable. It is also important that recommendations have a novelty degree~\cite{Gravino2019May}: to recommend \emph{Star Wars VI (Return of the Jedi)} to a Star Wars fan can be very accurate but it is not surprising or ``novel'' to him or her. Remarkable current research in \ac{RS} points to ``reliable'' recommendations~\cite{bobadilla2018reliability}: when an \ac{RS} recommends a restaurant by giving it five stars, a person is probably not entirely convinced about the rating. Some \ac{RS} services provide additional information that allows us to infer a ``reliability'' for the rating; this information usually consists of the number of people who have voted for the restaurant. Almost everyone prefers a restaurant with an average vote of 4 stars and 1200 ratings than a restaurant with an average vote of 5 stars and only 5 ratings. \ac{RS} methods and models for reliability automatically provide a reliability value that is associated with each recommendation.

Providing accurate reliability values that are associated with predictions is an important goal in the \ac{RS} field for the following reasons:

\begin{enumerate}[label=\alph*)]
    \item This approach resembles human semantics: ``I recommend this film because it is truly good, but I am not sure if you like war movies'' (high prediction, low reliability); ``I know you very well, and I am sure you will love these shoes'' (high prediction, high reliability); ``I definitely do not recommend this restaurant'' (low prediction, high reliability); ``I do not think that you are going to like this song'' (low prediction, low reliability).
    \item The extra ``reliability'' information enables new ways of showing recommendations to the users: from bidimensional plots (prediction, accuracy) to natural language, as demonstrated in the previous examples.
    \item People feel more comfortable when they use fuzzy concepts (you will like it ``a lot''; you ``probably'' will not like it) rather than absolute concepts (you will like it or will not like it).
    \item Reliability values can be used to improve the accuracy of the results instead of improving explanations. As we will show later, some related works address this problem. Accuracy improvements are mainly achieved by filtering low reliability recommendations.
    \item Reliability information is a powerful tool to improve users’ trust in a RS: categorical mistakes (such as erroneous five-star recommendations) generate distrust on the part of the users who receive the recommendations. Conversely, people tolerate errors from noncategorical information. Reliability values are adequate information to modulate the probability of recommendations (it is ``very likely'' that you like...).
    \item Reliability of predictions can be employed to properly handle a cold start problem. If all the predictions for a user or item have associated low reliability values, it can be considered a cold start problem, and recommendation methods that are designed ad hoc for cold start situations could be applied.
\end{enumerate}

The kernel of an \ac{RS} is its filtering approach; mainly, recommendations can be made based on demographic~\cite{AlShamri2016May}, content~\cite{Zamani2018Oct}, social~\cite{Rezvanian2019}, context~\cite{Villegas2018Jan} and collaborative information. Due to their accurate results, a \ac{CF} based \ac{RS}~\cite{Saquib2017} is the most extended \ac{RS}. Commercial \ac{RS} architectures usually implement hybrid approaches~\cite{Ignatev2018Nov} that include \ac{CF} and other filtering sources. Research on reliability has been focused on the \ac{CF} context, and methods and models have been published to provide reliability values that are mainly based on both the \ac{KNN} approach and the \ac{MF} approach. A \ac{KNN}-based framework for providing reliability information was proposed in~\cite{Hernando2013Jan}, where a specific implementation of the framework is explained. The key idea is that positive and negative factors are analyzed to obtain reliabilities: the greater are the positive factors, the higher are the reliability values; and the greater are the negative factors, the lower are the reliability values. A completely different \ac{KNN} approach to obtaining reliabilities is proposed in~\cite{Ahmadian2019Jul}, where low-quality rating profiles of users are enhanced by adding reliable votes. To estimate the confidence level of each prediction,~\cite{10.1145/2043932.2043956} provides a full probability distribution of the item ratings rather than only a single score.

Trust information is a relevant concept that is related to reliability since both are usually correlated. In~\cite{Gohari2018Jan}, the authors propose a confidence-based \ac{RS} approach where trust and certainty are combined. The paper addresses the uncertainty that is derived from different rating behaviors of users on the same scale (e.g., 1 to 5 stars), and trust is applied to the opinions of the users. The use of ontologies and fuzzy linguistic models to represent users’ trust in \acp{RS} is addressed in~\cite{Martinez-Cruz2015Aug}, where users’ trust is employed in \ac{RS} filtering rather than traditional similarity based on ratings. A Boltzmann machine learning model that is fed with trust information and users’ preferences is proposed in~\cite{Wu2019Sep}, and the model merges both inputs to obtain improved recommendations. Trust-aware information was utilized in~\cite{Azadjalal2017Jan} to enrich the prediction process and obtain a ``confidence'' value for a recommendation. Trust-aware information~\cite{10.1145/2492517.2500276} has been employed in other works to obtain reliability values for predictions and recommendations. Based on a reliability-based method,~\cite{Moradi2015Nov} improves the \ac{RS} accuracy by providing a dynamic mechanism to construct trust networks of the users. It is particularly important to be aware that reliability values can be applied to improve the accuracy results by selecting the recommendations with the highest reliability values~\cite{Hernando2013Jan}. Some trust-aware confidence measures for rating predictions were proposed in~\cite{Moradi2015Oct}. Moreover, beyond accuracy, the entropy concept~\cite{Wang2015Aug} can be applied to an \ac{RS} to create a reliability measure. The initialization of \ac{MF} is improved in~\cite{Deng2016Feb} by using social trust information and a deep learning model.

The concept of reliability or confidence has also been recently addressed by~\cite{Mesas2020Feb} and~\cite{Margaris2020Nov}. In the first paper, the authors focus on several methods to embed awareness in an \ac{RS} to determine whether each item should be suggested. Similar to our proposed method, the authors claim that confidence values can be used to filter the most reliable suggestions, which leads to an accuracy increase and a coverage decrease. The second paper~\cite{Margaris2020Nov} introduces the reliability concept in the review-to-rating conversion field, that is, the reliability in the conversion of textual reviews to implicit ratings. An aggregation of reliability measures was proposed in~\cite{Ahmadian2019Jul}, where user-based reliability, rating profile reliability and item-based reliability were combined. By using this information, the authors obtain improved recommendations. Nevertheless, this solution is a memory-based solution, and thus, it has important drawbacks compared to our proposed model-based approach. Analogously, a confidence measure and a similarity confidence coefficient for link predictions are proposed in~\cite{Su2019Aug} in the context of the \ac{KNN} memory-based method. Another memory-based approach~\cite{Fan2018} utilizes the reliability concept to improve traditional Pearson and cosine similarity measures. Previously, authors created the ``credibility of a rating between users'' based on the users’ ratings on common items. \acp{RS} can also be improved by using context-aware prefiltering. Contextual recommendations were made in~\cite{Ramirez-Garcia2015Jun} using traditional \ac{CF} and fuzzy rules to select items that are suitable for the specific situations of the users. In the same way, \ac{CF} contextual postfiltering is performed in~\cite{Ramirez-Garcia2014Mar} to cast recommendations of restaurants with the aim of disseminating information about products and services. Moreover, an increase in effectiveness in advertising campaigns has been reached by combining evolutionary computation and fuzzy logic~\cite{Madera2017Nov}. Additionally, online advertising recommendations have been improved using confidence information~\cite{Nguyen-Thanh2019Sep}, which outperforms some current reinforcement learning schemes. Both~\cite{Liang2021Jan} and~\cite{Xu2019Jun} are research papers that use an upper bound of confidence to improve recommendations. The former work, \cite{Liang2021Jan}, has been tested in event-based social networks and synthetic datasets. On the other hand, in~\cite{Xu2019Jun}, a user’s confidence and time context algorithm are proposed, where the user’s behavior temporal information is exploited to improve the recommendation accuracy.

Despite its success, the previously mentioned research has two main drawbacks in the current \ac{RS} reliability issue. First, the \ac{KNN} approach is not applied due to its lack of accuracy and scalability. Conversely, our proposed method is not based on \ac{KNN} but is based on the \ac{MF} model, which is now a \ac{CF} standard~\cite{Zhang2019Nov}. Second, social information is only available in a small subset of existing \acp{RS}, so social-based trust-aware approaches that are oriented to obtain reliability values cannot be considered universal solutions. Conversely, our proposed approach only uses the ratings matrix that contains the preferences of a set of users over a set of items, which is basic information for \ac{CF}-based \acp{RS}. Since the proposed method does not rely on the \ac{KNN} method, where identifying strategies for designing reliability measures are more obvious, it utilizes \ac{MF} techniques and provides a brand-new \ac{MF}-based approach to address \ac{RS} reliabilities.

We will employ remarkable research in the \ac{MF} field that has been applied to an \ac{RS} in this paper as baselines. \ac{PMF}~\cite{mnih2008probabilistic} and Biased \ac{MF}~\cite{Koren2009Aug} are classic \ac{MF} implementations. \ac{NMF}~\cite{NIPS2000_1861} avoids negative latent factor values and facilitates the assignment of semantic meanings to them. \ac{BNMF}~\cite{Hernando2016Apr} provides an understandable probabilistic meaning to the latent factors, which lie within the range $\left [ 0, 1 \right ]$, and group users that share the same tastes. An improved \ac{CF} Singular Value Decomposition (SVD++) is presented in~\cite{Koren2008}. The \ac{URP} model~\cite{Marlin2003} produces complete user rating profiles; each item rating is assigned by selecting a user attitude for the item. A two-level \ac{MF} semantic architecture is designed in~\cite{zhu2018assigning}, which allows us to add reliabilities for any \ac{MF}-based \ac{CF} system as an add-on. We will use this method to obtain reliability values from the previously described \ac{MF} implementations; it serves as a base for this paper’s baselines.

One of the reasons that explains why the problem of reliability in \ac{RS} has not been completely addressed is the lack of quality measures of reliability. Accuracy quality can be measured using a complete set of prediction and recommendation quality measures (\ac{MAE}, \ac{RMSE}, precision, recall, F1, etc.); reliability does not have analogous quality values. The underlying idea is that \textit{``the more suitable reliability is, the better accuracy results will provide when applied: predictions with higher reliabilities should provide more accurate (less error) results, whereas we expect higher prediction errors on low reliability recommended items''}~\cite{zhu2018assigning}. This is a recurrent concept that is addressed in several papers: \textit{``The most common measurement of confidence is the probability that the predicted value is indeed true''}~\cite{Shani2011}, \textit{``This reliability measure is based on the usual notion that the more reliable a prediction, the less liable to be wrong''}~\cite{Hernando2013Jan}, and \textit{``Evaluations of recommenders for this task must evaluate the success of high-confidence recommendations, and perhaps consider the opportunity costs of excessively low confidence''}~\cite{Herlocker2004Jan}. Using these concepts, the reliability framework proposed in~\cite{Hernando2013Jan} provides plots that show the prediction errors versus reliability inverse correlation. Based on confidence curve analysis, \cite{Mazurowski2013Aug} proposed a method to estimate the reliability quality of reliability measures. The method tests the confidence curve by checking that its first value is higher than the last value (error decreases when reliability increases). The weak point of~\cite{Hernando2013Jan} is that it does not test all the confidence curve trends. In~\cite{bobadilla2018reliability}, the \ac{RPI}, which is a quality measure of reliability that tests the quality of reliabilities for predictions, is proposed. This quality measure returns a simple value, similar to customary accuracy quality measures, and scores the quality of a reliability measure for an \ac{RS}.

The proposed model in this paper obtains the reliability value of each prediction and recommendation using three concepts: the machine learning classification concept, Bernoulli distribution and \ac{MF} model. Instead of designing a regression model, we have chosen a classification approach, so the classes to classify correspond to the finite set of possible scores that can be assigned as rates to the items of the RS, e.g., $\left \{1,2,3,4,5\right \}$ stars in the MovieLens dataset~\cite{harper2015movielens}. In this way, our model returns the probabilities that a user will assign each of the possible scores to an item. In our MovieLens example, the model may return $\left [0.05, 0.05, 0.1, 0.6, 0.2\right ]$, which means that the score $4$ (probability $0.6$) is the best prediction choice (for a given $\left \langle \mathrm{user}, \mathrm{item} \right \rangle$ pair). The classification model in the \ac{RS} \ac{MF} context is an innovative approach that is proposed in this paper: current \ac{MF} implementations return a regression value for each prediction, e.g., $4.3$. To design the explained classification-based \ac{MF} approach, we set $D$ individual classification tasks independent from each other, where $D$ is the number of possible scores in the dataset ($D = 5$ in MovieLens). In this way, we have $D$ separated classification processes. Each classification task for a given score focuses on the dichotomy that this score may or may not correspond to the score evaluated for each $\left \langle \mathrm{user}, \mathrm{item} \right \rangle$ pair. This behavior can be modeled using the Bernoulli distribution, that is, the discrete probability distribution of a random variable that takes the value $1$ with probability $p$ and the value $0$ with probability $1-p$. The formalization of the classification-based \ac{MF} approach by using the Bernoulli distribution is detailed in~\cref{sec:bemf}. We have named this proposed method \ac{BeMF} in allusion to the Bernoulli distribution and the \ac{MF} model.

\begin{figure}[ht]
    \centering
    \includegraphics[width=\textwidth]{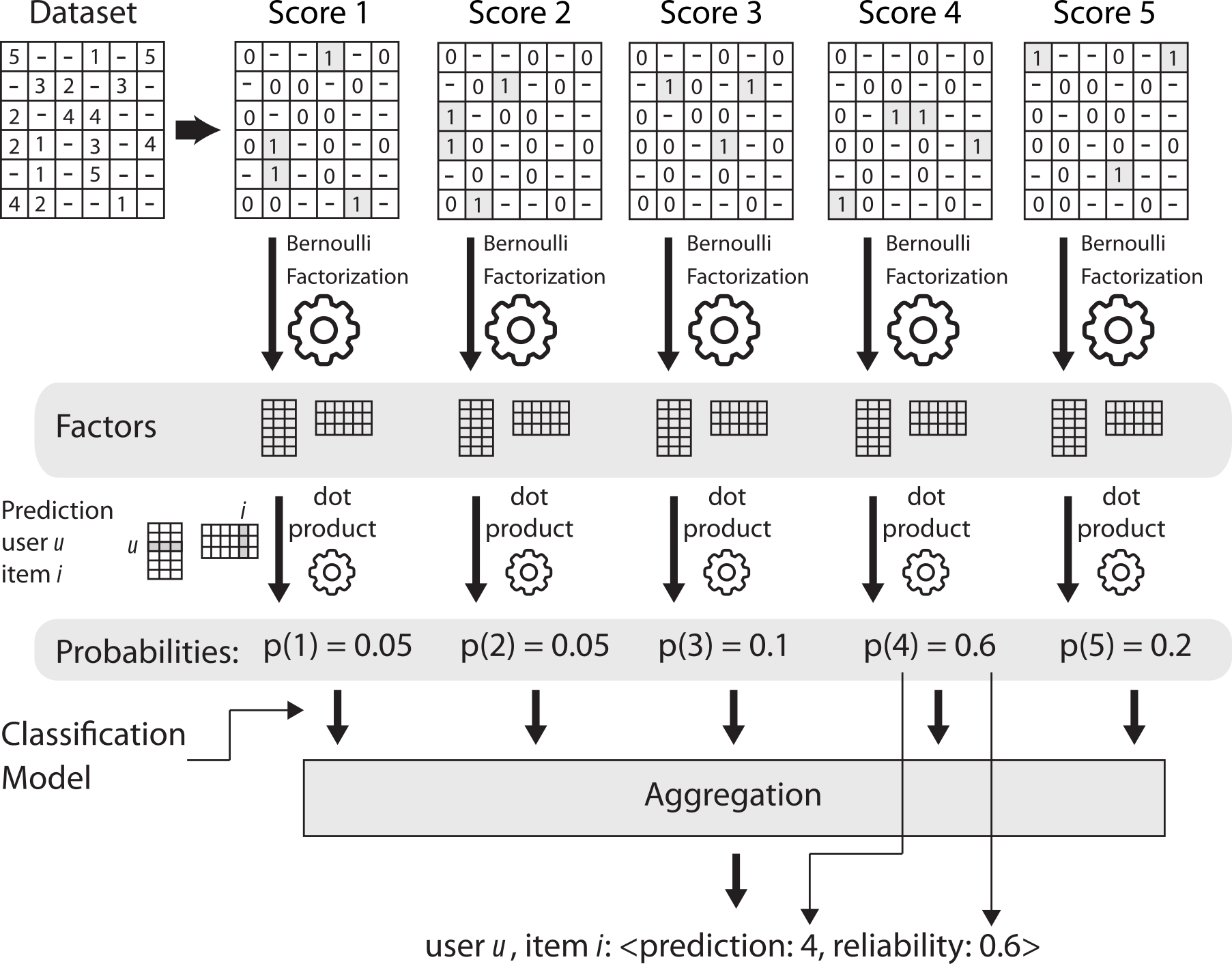}
    \caption{Architecture of the proposed \ac{BeMF} model}\label{fig:architecture}
\end{figure}

\Cref{fig:architecture} shows the main ideas of the proposed \ac{BeMF} model using a graphical example: the dataset displayed as a rating matrix is split into $D = 5$ matrices, which correspond to the possible scores (score 1, score 2, \ldots, score 5). Each matrix contains the basic information that is related to each corresponding score: if the rating corresponds to the actual score, we assign the code ``1''; if the rating does not correspond to the score, we assign the code ``0''; otherwise, the nonrated mark ``-'' is assigned. This set of $D = 5$ separated matrices feeds $D = 5$ independent Bernoulli factorizations, and as usual, each factorization generates its user’s latent factors matrix and its item’s latent factor matrix (gray row named \textit{``Factors''}). From each pair of latent matrices, we can obtain predictions that utilize the inner product of the chosen user and item factors. Note that each prediction result does not correspond to the regression value of the rating, as is customary, but corresponds to the probability that the score will be classified as the correct score (gray row named \textit{``Probabilities''}). The summation of these probabilities is equal to 1 since the \ac{BeMF} model returns the probability distribution of the rating of user $u$ to item $i$. The last step is to aggregate the $D = 5$ resulting probabilities to establish a classification ``winner''. Usually, the aggregation approach selects the score that is associated with the maximum probability. Note that we refer to Bernoulli factorization for the $D$ \ac{MF} parallel processes that are based on Bernoulli distribution, whereas we refer to the \ac{BeMF} model for the complete architecture.

The proposed model provides an innovative approach to obtain reliability values in a \ac{CF} \ac{RS}. The main advantages of the model are listed as follows:

\begin{enumerate}[label=\alph*)]
    \item \ac{BeMF} can be considered universal since it can be applied to a \ac{CF} \ac{RS}, which is are just based on ratings. Particularly, \ac{BeMF} does not use social information, which is important since social information is only available in a reduced set of \ac{CF} \acp{RS}. A variety of papers rely on social information and the trust and distrust of the associated information~\cite{Azadjalal2017Jan,Moradi2015Nov,Moradi2015Oct,10.1145/2492517.2500276}, but they cannot be applied to the most popular \ac{RS} datasets.
    
    \item \ac{BeMF} is a model-based approach, and therefore, it is not limited to small- or medium-sized \acp{RS}, as discussed in papers on memory-based \ac{KNN}~\cite{Ahmadian2019Jul,Hernando2013Jan,10.1145/2043932.2043956}. In this way, the proposed approach is scalable, whereas the memory-based methods are not scalable.
    
    \item The \ac{BeMF} model is based on classification~\cite{Batmaz2019Jun,Bobadilla2020Jan} rather than the usual regression approaches~\cite{Hernando2016Apr,Koren2008,Koren2009Aug,NIPS2000_1861,Marlin2003,mnih2008probabilistic}. The machine learning classification approach provides more information than the regression approach; specifically, it returns a set of $D$ probabilities. The $D$ probabilities can be aggregated to obtain the expected reliability result. This context is richer than the isolated resultant value that is provided by the regression models.
    
    \item The \ac{BeMF} machine learning model is based on the Bernoulli distribution because it adequately adapts to the binary expected result for each of the $D$ possible scores. The key concept is the simplification that we make by performing $D$ simpler and more precise processes instead of the traditional complex regression task. To the best of our knowledge, a Bernoulli classification-based model has never been published in the reliability \ac{RS} field. Additionally, the Bernoulli distribution directly provides, in a natural way, the expected probability that the prediction is true, that is, reliability.
    
    \item Despite its simplicity, the use of the Bernoulli distribution endows \ac{BeMF} with considerable flexibility. Eventually, the output of the \ac{BeMF} model for user $u$ and item $i$ can be understood as a vector $(p^1_{u,i}, \ldots, p^D_{u,i})$ of probabilities, where $p^s_{u,i}$ is the probability that $u$ assigns the $s$-th score (for example, $s$ stars) to item $i$. This distribution is a discrete probability distribution on the set of possible ratings. By using the underlying Bernoulli distribution, \ac{BeMF} can actually model any discrete probability on the set of ratings or it fits the hyperparameters of a categorical distribution. Thus, it is completely general from a probabilistic point of view. The difference between \ac{BeMF} and a model that is based on a categorical distribution is the training method: using the Bernoulli distribution for each score, we assume that the ratings are independent. For this reason, each probability $p_s$ is computed only based on the appearances of the ratings $s$ on the dataset. This process accelerates the training process and prevents overfitting.
\end{enumerate}

This paper’s hypothesis claims that an \ac{MF} model can be designed to provide the reliability of each \ac{RS} recommendation by using a classification-based approach and setting its learning stage in the Bernoulli distribution. Our contribution is the proposal and validation of an \ac{RS} model that is capable of providing not only recommendations using a classification-based approach but also the reliability of these recommendations. Regarding the originality of our contribution, this is an innovative approach that provides explicit reliability values to \ac{RS} users or improves the accuracy by selecting recommendations with the highest reliability values. Although reliability has been employed to improve accuracy, it has not become a common practice. Additionally, the reliability measures that are provided by this machine learning model are intrinsically linked to the model and do not depend on external methods or extended architectures as existing solutions.

The remainder of the paper has been structured as follows: In \cref{sec:bemf}, the mathematical foundations of the proposed method are explained, and the resulting \ac{BeMF} algorithm is provided. \Cref{sec:evaluation} shows the designed experiments and their results and discussions. \Cref{sec:conclusions} contains the main conclusions of the paper and outlines some future works.

\section{Proposed model}\label{sec:bemf}

Following the standard framework of \acp{CF}, we assume that we have $N$ users that are evaluating $M$ different items with a discrete set of possible scores $\mathcal{S} = \left\{s_1, \ldots, s_D\right\}$ (typically $\mathcal{S} = \left\{1, \ldots, 5\right\}$ as in the MovieLens dataset~\cite{harper2015movielens}). These ratings are collected in the rating matrix $R = (R_{u,i})$, where $R_{u,i} = s_\alpha$ if user $u \in [1, N]$ has assigned item $i \in [1, M]$ the score $s_\alpha \in \mathcal{S}$ and $R_{u,i} = \bullet$ if user $u$ has not rated item $i$. 

From this rating matrix, we generate $D$ distinct matrices $R^{s_1}, \ldots, R^{s_D}$, which correspond to the possible scores that can be assigned to the items. In this way, fixed $s \in \mathcal{S}$; the matrix $R^{s} = (R^s_{u,i})$ is a (sparse) matrix such that $R^s_{u,i} = 1$ if user $u$ voted item $i$ with exactly score $s$; $R^s_{u,i} = 0$ if user $u$ voted item $i$ but with a different score from $s$ and $R^s_{u,i} = \bullet$ if $u$ did not rate item $i$.

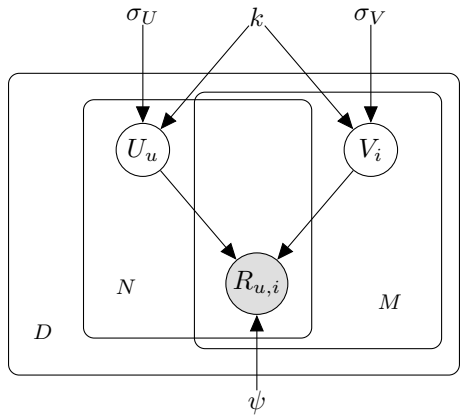
\begin{figure}[ht]
  \centering
  \begin{tikzpicture}
    \node[const] (sr) {$\psi$};
    \node[obs, above=of sr] (r) {$R_{u,i}$};
   
    \node[latent, above=of r, xshift=-1.5cm] (q) {$U_u$};
    \node[const, above=1.3cm of q] (sq) {$\sigma_U$};

    \node[const, above=3cm of r] (k) {$k$};

    \node[latent, above=of r, xshift=1.5cm] (p) {$V_i$};
    \node[const, above=1.3cm of p] (sp) {$\sigma_V$};

    \tikzset{plate caption/.style={caption, below right=-.4cm and -.4cm of #1.south east}}
    \plate [inner sep=.40cm] {rp} {(r)(p)} {$M$};
   
    \tikzset{plate caption/.style={caption, below left=-.6cm and -.4cm of #1.south west}}
    \plate [inner sep=.3cm] {rq} {(r)(q)} {$N$};
    
    \tikzset{plate caption/.style={caption, below left=0cm and 0.7cm of #1.south west}}
    \plate [inner sep=.50cm, xshift=0.3cm, yshift=0.15cm] {rq} {(q)(p)(r)} {$D$};

    \edge {p,q} {r}
    \edge {sp} {p}
    \edge {sq} {q}
    \edge {sr} {r}
    \edge {k} {p}
    \edge {k} {q}

  \end{tikzpicture}
  \caption{Plate diagram of \ac{BeMF} model.}
  \label{fig:plate}
\end{figure}

Our model will attempt to fit the matrices $R^{s_1}, \ldots, R^{s_D}$ by performing $D$ parallel matrix factorizations. The factors of each of the values will be independent, so the \ac{BeMF} model is the juxtaposition of $D$ different individual Bernoulli factorizations that fit binary matrices. \Cref{fig:plate} shows the plate diagram of the model, which is composed of $D$ different factorizations (one factorization for each possible score), each of which is composed of $N$ variables $U_u$ (latent factors of each user) and $M$ variables $V_i$ (latent factors of each item). The hyperparameters of the model are two positive real values $\sigma_U, \sigma_V > 0$: an integer $k > 0$ (number of hidden factors) and a smooth logistic-like function $\psi$.

The problem of predicting the missing values of the matrix $R^{s}$ is not a regression problem, as is customary in \ac{RS} literature. The matrix $R^s$ is a binary matrix, and each known value $R_{u,i}^s$ has a very specific meaning, namely, whether user $u$ considered that item $i$ deserves rate $s$. This problem is intrinsically a classification problem; in the simplest case it is a binary classification problem.

For this reason, the proposed system models the decision $R^s_{u,i}$, as a random variable of Bernoulli type with the probability of success $0 \leq p_{u,i}^s \leq 1$. This approach is compatible with the known fact that deciding to rate an item is not deterministic and depends on some stochastic psychological process that may vary from day to day. Hence, $p_{u,i}^s$ only measures the affinity of $u$ for item $i$ (more precisely, the reliability in the prediction, c.f., \cref{sec:bemf:model}). In this way, the higher is the value $p_{u,i}^s$, the more likely $u$ rates $i$ with score $s$, independent of imponderable factors.

Regarding how the value $p_{u,i}^s$ is estimated, which is consistent with the usual assumptions in \ac{RS} models, we will assume that the probability $p_{u,i}^s$ is given as the inner product $U_{u}^s \cdot V_{i}^s$ of some $k$-dimensional hidden user factors vector $U_u^s = (U_{u,1}^s, \ldots, U_{u,k}^s)$ and hidden item factors vector $V_i^s = (V_{i,1}^s, \ldots, V_{i,k}^s)$. However, the inner product $U_{u}^s \cdot V_{i}^s$  may attain any real value, so it is not a suitable quantity for $p_{u,i}^s$. To normalize the product $U_{u}^s \cdot V_{i}^s$, we use a logistic-like smooth function $\psi: \mathbb{R} \to [0,1]$ in such a way that $0 \leq \psi(U_{u}^s\cdot V_{i}^s) \leq 1$, and thus, we may consider $p_{u,v}^s = \psi(U_{u}^s\cdot V_{i}^s)$. In contrast with other \ac{MF} models, for each user and item, there is not a single instance of user and item factors but $D$ different vectors of factors, one vector of factor for each possible rating.

\subsection{Bernoulli factorization}\label{sec:bemf:factorization}

In this section, we describe the mathematical formulation of the factorization model for each score. Hence, throughout this section, we fix $s \in \mathcal{S}$ as a possible score. To shorten the notation, we denote $R = R^s$ as the vote matrix, which corresponds to the score $s$. This matrix is a sparse binary matrix.

First, we fix the logistic-like function $\psi: \mathbb{R} \to [0,1]$, i.e., a smooth increasing function with $\psi(x) \to 0$ when  $x \to -\infty$ and $\psi(x) \to 1$ when $x \to \infty$. This hyperparameter will have the role of an activation function that will translate the inner product of the factors into a probability.

Fix user $u$ and item $i$. The underlying probabilistic assumption of the Bernoulli factorization is that, given the latent vectors of user $U_u$ and item $V_i$ of dimension $k > 0$ (number of hidden factors), the rate $R_{u,i}$ is a Bernoulli distribution with the success probability $\psi(U_u \cdot V_i)$. Hence, the mass function of this random variable, $p(R_{u,i} | U_u, V_i)$, is given by

$$
    p(R_{u,i} | U_u, V_i) = \left\{\begin{matrix}\psi(U_uV_i) & \textrm{if } R_{u,i} = 1, \\ 1-\psi(U_uV_i) & \textrm{if } R_{u,i} = 0.\end{matrix}\right.
$$

From this formula, given sample $R = (R_{u,i})$, we obtain that the associated likelihood, $\cL(R | U, V)$, is

$$
    \cL(R | U, V) = \prod_{R_{u,i} \neq \bullet} p(R_{u,i} | U_u, V_i) = \left(\prod_{R_{u,i} = 1} \psi(U_uV_i)\right)\left(\prod_{R_{u,i} = 0} 1-\psi(U_uV_i)\right).
$$

In this way, the log-likelihood function, `$\cl(R | U, V) = \log \cL(R | U, V)$, is given by

$$
    \cl(R | U, V) =  \sum_{R_{u,i} = 1} \log(\psi(U_uV_i)) + \sum_{R_{u,i} = 0} \log(1-\psi(U_uV_i)).
$$

Now, as in \ac{PMF} \cite{mnih2008probabilistic}, let us assume spherical normal priors with a zero mean and standard deviation given by hyperparameters $\sigma_U, \sigma_V > 0$. In this case, we have the probability density functions

$$
    p(U_u) = \frac{1}{\sigma_U \sqrt{2\pi}} \exp\left(-\frac{||U_u||^2}{2\sigma^2}\right), \quad p(V_i) = \frac{1}{\sigma_V \sqrt{2\pi}} \exp\left(-\frac{||V_i||^2}{2\sigma^2}\right).
$$

Hence, their prior likelihoods, $\cL(U)$ and $\cL(V)$ for users and items, respectively, are

$$
    \cL(U) = \prod_{u=1}^N p(U_u) = \frac{1}{\sigma_U^N (2\pi)^{N/2}} \prod_{u=1}^N \exp\left(-\frac{||U_u||^2}{2\sigma^2_U}\right) = \frac{1}{\sigma_U^N (2\pi)^{N/2}}  \exp\left(-\frac{\sum_{u=1}^N ||U_u||^2}{2\sigma^2_U}\right),
$$
$$
    \cL(V) = \prod_{i=1}^M p(V_i) = \frac{1}{\sigma^M_V (2\pi)^{M/2}} \prod_{i=1}^M \exp\left(-\frac{||V_i||^2}{2\sigma_V^2}\right) = \frac{1}{\sigma_V^M (2\pi)^{M/2}}  \exp\left(-\frac{\sum_{i=1}^M ||V_i||^2}{2\sigma_V^2}\right).
$$

Analogously, their prior log-likelihoods are

$$
    \cl(U) = -\frac{1}{2\sigma_U^2}\sum_{u=1}^N ||U_u||^2 + C_U, \quad
    \cl(V) = -\frac{1}{2\sigma_V^2}\sum_{i=1}^M ||V_i||^2 + C_V.
$$

for some constants $C_U = -N\log(\sigma_U \sqrt{2\pi})$ and $C_V = -M\log(\sigma_V \sqrt{2\pi})$.

Moreover, the posterior likelihood, $\cL(R)$, is

$$
    \cL(R) = \cL(R | U, V) \cL(U) \cL(V).
$$

Therefore, the posterior log-likelihood is given by

\begin{align*}
\cl(R) &= \cl(R |U,V) + \cl(U) + \cl(V) \\
&= \sum_{R_{u,i} = 1} \log(\psi(U_uV_i)) + \sum_{R_{u,i} = 0} \log(1-\psi(U_uV_i)) \\ &\hspace{0.5cm}-\frac{1}{2\sigma_U^2}\sum_{u=1}^N ||U_u||^2 -\frac{1}{2\sigma_V^2}\sum_{i=1}^M ||V_i||^2 + C,
\end{align*}

where $C = C_U + C_V$ is a constant.

The maximum likelihood estimator is obtained by maximizing the posterior log-likelihood $\cl(R)$. For this purpose, the constant $C$ is irrelevant, so it can be dismantled. Setting $\eta_U = \frac{1}{\sigma_U^2}$ and $\eta_V = \frac{1}{\sigma_V^2}$, the maximization problem can be converted to the minimization of the cost function

$$
    \cF(U, V) = -\sum_{R_{u,i} = 1} \log(\psi(U_uV_i)) - \sum_{R_{u,i} = 0} \log(1-\psi(U_uV_i)) + \frac{\eta_U}{2}\sum_{u=1}^N ||U_u||^2 + \frac{\eta_V}{2}\sum_{i=1}^M ||V_i||^2.
$$

To optimize this cost function, we will use a standard gradient descent algorithm. Fixing user $u_0$ and an item $i_0$, denote the components of the latent vectors as $U_{u_0} = (U_{u_0, 1}, \ldots, U_{u_0, k})$ and $V_{i_0} = (V_{i_0, 1}, \ldots, V_{i_0, k})$. The directional derivatives of $\cF$ in the directions $U_{u_0, a}$ and $V_{i_0, b}$ are then given by

$$
    \frac{\partial  \cF}{\partial U_{u_0,a}} = -\sum_{\left\{i \,|\, R_{u_0,i} = 1\right\}} \frac{\psi'(U_{u_0}V_i)}{\psi(U_{u_0}V_i)}V_{i,a} + \sum_{\left\{i \,|\, R_{u_0,i} = 0\right\}} \frac{\psi'(U_{u_0}V_i)}{1-\psi(U_{u_0}V_i)}V_{i,a} + \eta_U U_{u_0,a},
$$
$$
    \frac{\partial  \cF}{\partial V_{i_0,b}} = -\sum_{\left\{u \,|\, R_{u,i_0} = 1\right\}} \frac{\psi'(U_{u}V_{i_0})}{\psi(U_{u}V_{i_0})}U_{u,b} + \sum_{\left\{u \,|\, R_{u,i_0} = 0\right\}} \frac{\psi'(U_{u}V_{i_0})}{1-\psi(U_{u}V_{i_0})}U_{u,b} + \eta_V V_{i_0,b}.
$$

For simplicity, let us set $\eta_U = \eta_V = \eta$. The associated gradient descent algorithm with step $\gamma > 0$ then updates the approximations at time $T$, which are denoted $U_u^{T}$ and $V_i^{T}$, to the approximations at time $T+1$ by the rule

$$
    U_u^{T+1} = U_u^{T} + \gamma \left(\sum_{\left\{i \,|\, R_{u,i} = 1\right\}} \frac{\psi'(U_{u}V_i)}{\psi(U_{u}V_i)}V_{i} - \sum_{\left\{i \,|\, R_{u,i} = 0\right\}} \frac{\psi'(U_{u}V_i)}{1-\psi(U_{u}V_i)}V_{i} - \eta U_{u}\right),
$$
$$
    V_i^{T+1} = V_i^{T} + \gamma \left(\sum_{\left\{u \,|\, R_{u,i} = 1\right\}} \frac{\psi'(U_{u}V_{i})}{\psi(U_{u}V_{i})}U_{u} - \sum_{\left\{u \,|\, R_{u,i} = 0\right\}} \frac{\psi'(U_{u}V_{i})}{1-\psi(U_{u}V_{i})}U_{u} - \eta V_{i}\right).
$$

\begin{ex}\label{ex:logit}
If we consider $\psi = \mathrm{logit}$ to be the logistic function, $\mathrm{logit}(x) = \frac{1}{1+e^{-x}}$, then we have that $\mathrm{logit}'(x) = \mathrm{logit}(x) (1-\mathrm{logit}(x))$. Hence, the update rule reads

\begin{align*}
    U_u^{T+1} = U_u^{T} + \gamma &\left(\sum_{\left\{i \,|\, R_{u,i} = 1\right\}} \frac{\mathrm{logit}(U_{u}V_i)(1-\mathrm{logit}(U_{u}V_i))}{\mathrm{logit}(U_{u}V_i)}V_{i} \right. \\
    &\left. \quad- \sum_{\left\{i \,|\, R_{u,i} = 0\right\}} \frac{\mathrm{logit}(U_{u}V_i)(1-\mathrm{logit}(U_{u}V_i))}{1-\mathrm{logit}(U_{u}V_i)}V_{i} - \eta U_{u}\right),
\end{align*}
\begin{align*}
    V_i^{T+1} = V_i^{T} + \gamma &\left(\sum_{\left\{u \,|\, R_{u,i} = 1\right\}} \frac{\mathrm{logit}(U_{u}V_i)(1-\mathrm{logit}(U_{u}V_i))}{\mathrm{logit}(U_{u}V_{i})}U_{u}\right. \\
    &\left. \quad-  \sum_{\left\{u \,|\, R_{u,i} = 0\right\}} \frac{\mathrm{logit}(U_{u}V_i)(1-\mathrm{logit}(U_{u}V_i))}{1-\mathrm{logit}(U_{u}V_{i})}U_{u} - \eta V_{i}\right).
\end{align*}

These update rules can be simplified as

$$
    U_u^{T+1} = U_u^{T} + \gamma \left(\sum_{\left\{i \,|\, R_{u,i} = 1\right\}} (1-\mathrm{logit}(U_{u}V_i))V_{i} - \sum_{\left\{i \,|\, R_{u,i} = 0\right\}} \mathrm{logit}(U_{u}V_i)V_{i} - \eta U_{u}\right),
$$
$$
    V_i^{T+1} = V_i^{T} + \gamma \left(\sum_{\left\{u \,|\, R_{u,i} = 1\right\}} (1-\mathrm{logit}(U_{u}V_i))U_{u} - \sum_{\left\{u \,|\, R_{u,i} = 0\right\}} \mathrm{logit}(U_{u}V_i)U_{u} - \eta V_{i}\right).
$$
\end{ex}

\subsection{\ac{BeMF} model}
\label{sec:bemf:model}

By performing Bernoulli factorization on each possible score, we obtain a collection of user and item factor pairs $(U^{s_1}, V^{s_1}), \ldots, (U^{s_D}, V^{s_D})$ for the scores $\mathcal{S} = \left\{s_1, \ldots, s_D\right\}$. We can gather all this information to obtain the final output $\Phi$. For each user $u$ and item $i$, the output is the $D$-dimensional vector

$$
    \Phi(u,i) = \frac{1}{\sum\limits_{\alpha = 1}^s \psi(U_u^{s_\alpha} V_i^{s_\alpha})} \left(\psi(U_u^{s_1} V_i^{s_1}), \ldots, \psi(U_u^{s_D} V_i^{s_D})\right).
$$

This vector is vector $\Phi(u,i) = (p^1_{u,i}, \ldots, p^D_{u,i})$ with $0 \leq p^\alpha_{u,i} \leq 1$ and $\sum_\alpha p^\alpha_{u,i} = 1$. The value $p^{\alpha}_{u,i}$ may be interpreted as the probability that user $u$ assigns score $s_{\alpha}$ to item $i$. In this way, $p^\alpha_{u,i}$ is the reliability that we have in predicting $s_\alpha$. From this vector, if $\alpha_0 = \textrm{argmax}_\alpha\,\, p^\alpha_{u,i}$ is the mode of this probability distribution, we obtain

\begin{itemize}
    \item The prediction is given by $\hat{R}_{u,i} = s_{\alpha_0}$.
    \item The reliability of the prediction is given by $\rho_{u,i} = p^{\alpha_0}_{u,i}$.
\end{itemize}

Moreover, we can fix the threshold $\vartheta > 0$ (required prediction reliability), and we set $\hat{R}_{u,i} = \bullet$ (i.e., the prediction is unreliable if $\rho_{u,i} < \vartheta$). In this way, we filter unreliable predictions.

\subsection{\ac{BeMF} algorithm}
\label{sec:bemf:algorithm}

Algorithm\ref{alg:bemf} contains the pseudocode for the training process of the \ac{BeMF} model using the logistic function ($\psi = \mathrm{logit}$) as the activation function, as shown in example~\ref{ex:logit}. The algorithm receives as inputs the sparse rating matrix ($R$), number of latent factors ($k$) and hyperparameters that are required by gradient descent optimization: learning rate ($\gamma$), regularization ($\eta$) and number of iterations ($m$). The algorithm returns as output two matrices that contain the learned latent factors: $U$ contains the latent factors for each score $s$, user $u$ and factor $f$, and $V$ contains the latent factors for each score $s$, item $i$ and factor $f$. These matrices can be used to compute both predictions and reliability, as described in~\cref{sec:bemf:model}. Note that to reduce the algorithm’s processing time, the user update loop (lines 5-17) and item update loop (lines 20-32) can be executed in parallel for each user and item, respectively.

\begin{algorithm}[H]
	\label{alg:bemf}
	\DontPrintSemicolon
	\LinesNumbered
	
    \SetKwInOut{Input}{input}
    \SetKwInOut{Output}{output}

    \Input{$R, k, \gamma, \eta, m, \mathcal{S}$}
	\Output{$\mathbf{U}, \mathbf{V}$}

	Initialize $\mathbf{U}\leftarrow U(0,1), \mathbf{V}\leftarrow U(0,1)$\;

	\Repeat{$m$ iterations} {
    	\For{each possible score $s \in \mathcal{S} = \left \{ s_1, \ldots, s_D \right \}$} {
    	    \For{each user $u$}{
    	        Initialize $\Delta$ to 0\;
        	    \For{each item $i$ rated by user $u$: $R_{u,i}$}{
            	    \For{each $f \in \{1, \ldots, k\}$}{
            	        \uIf{$R_{u,i}=s$}{
                            $\Delta_f^s \leftarrow \Delta_f^s + (1-\mathrm{logit}({U}_u^{s} \cdot V_i^{s})) \cdot {V}_{i,f}^{s}$ \;
                        }
                        \Else{
                            $\Delta_f^s \leftarrow \Delta_f^s - \mathrm{logit}({U}_u^{s} \cdot {V}_i^{s} ) \cdot {V}_{i,f}^{s}$ \;
                        }
            	    }
            	}
            	\For{each $f \in \{1, \ldots, k\}$}{
        	        $U_{u,f}^{s} \leftarrow U_{u,f}^{s} + \gamma\cdot (\Delta_f^s - \eta U_{u,f}^s)$\;
        	    }
    	    }
        	\For{each item $u$}{
        	    Initialize $\Theta$ to 0\;
            	\For{each user $u$ that rated item $i$: $R_{u,i}$}{
            	    \For{each $f \in \{1, \ldots, k\}$}{
            	        \uIf{$R_{u,i}=s$}{
                            $\Theta_f^s \leftarrow \Theta_f^s + (1-\mathrm{logit}({U}_u^{s} \cdot {V}_i^{s})) \cdot {U}_{u,f}^{s}$ \;
                        }
                        \Else{
                            $\Theta_f^s \leftarrow \Theta_f^s - \mathrm{logit}({U}_u^{s} \cdot {V}_i^{s}) \cdot {U}_{u,f}^{s}$ \;
                        }
            	    }
            	}
            	
            	\For{each $f \in \{1, \ldots, k\}$}{
            	    $V_{i,f}^{s} \leftarrow V_{i,f}^{s} + \gamma\cdot (\Theta_f^s - \eta V_{i,f}^s)$\;
            	}
        	}
        }
    }
 
	\caption{\ac{BeMF} model fitting algorithm}
\end{algorithm}

\subsection{Running example}\label{sec:bemf:example}

This section presents a running example of the \ac{BeMF} model using the dataset with 4 users and 6 items shown in~\cref{tab:example-dataset}. To reduce the extension of the running example and make it more readable, we set the possible scores to `like' (\textcolor{Green}{\faThumbsOUp}) and `dislike' (\textcolor{Red}{\faThumbsODown}). Note that despite the simplicity of this set of possible scores, it is a real set of possible scores that is employed in commercial services such as YouTube, Tinder or Steam.

\begin{table}[h]
\begin{tabularx}{\textwidth}{ 
  | >{\centering\arraybackslash}X 
  | >{\centering\arraybackslash}X 
  | >{\centering\arraybackslash}X 
  | >{\centering\arraybackslash}X
  | >{\centering\arraybackslash}X 
  | >{\centering\arraybackslash}X 
  | >{\centering\arraybackslash}X | }
    \hline 
    $R_{u,i}$ & $i_1$                           & $i_2$                           & $i_3$                           & $i_4$                           & $i_5$                           & $i_6$                           \\ 
    \hline
    $u_1$     & \textcolor{Red}{\faThumbsODown} & \textcolor{Green}{\faThumbsOUp} &                                 & \textcolor{Green}{\faThumbsOUp} &                                 &                                 \\ 
    \hline
    $u_2$     & \textcolor{Green}{\faThumbsOUp} &                                 & \textcolor{Green}{\faThumbsOUp} & \textcolor{Red}{\faThumbsODown} & \textcolor{Red}{\faThumbsODown} &                                 \\ 
    \hline
    $u_3$     &                                 & \textcolor{Green}{\faThumbsOUp} & \textcolor{Green}{\faThumbsOUp} &                                 &                                 & \textcolor{Red}{\faThumbsODown} \\ 
    \hline
    $u_4$     & \textcolor{Red}{\faThumbsODown} &                                 & \textcolor{Red}{\faThumbsODown} &                                 & \textcolor{Green}{\faThumbsOUp} & \textcolor{Green}{\faThumbsOUp} \\ 
    \hline
\end{tabularx}
\caption{Rating matrix used in the running example. Possible scores are set to like (\textcolor{Green}{\faThumbsOUp}) and dislike (\textcolor{Red}{\faThumbsODown}).}
\label{tab:example-dataset}
\end{table}

The first step is to split the rating matrix into as many matrices as the number of possible scores. In this case, we have two possible scores, so we split the rating matrix of~\cref{tab:example-dataset} into a matrix that encodes the `like' ratings (\cref{tab:example-dataset-split:like}) and a matrix that encodes the `dislike' ratings (\cref{tab:example-dataset-split:dislike}). Recall that the absence of a rating must not be encoded into these matrices.

\begin{table}[h]
\centering
\begin{subtable}{.48\textwidth}
    \centering
    \begin{tabularx}{\textwidth}{ 
      | >{\centering\arraybackslash}X 
      | >{\centering\arraybackslash}X 
      | >{\centering\arraybackslash}X 
      | >{\centering\arraybackslash}X
      | >{\centering\arraybackslash}X 
      | >{\centering\arraybackslash}X 
      | >{\centering\arraybackslash}X | }
        \hline 
        $R_{u,i}^{\text{\faThumbsOUp}}$ & $i_1$ & $i_2$ & $i_3$ & $i_4$ & $i_5$ & $i_6$ \\ 
        \hline
        $u_1$                           & 0       & 1     &       & 1    &        &       \\ 
        \hline
        $u_2$                           & 1       &       & 1     & 0    & 0      &       \\ 
        \hline
        $u_3$                           &         & 1     & 1     &      &        & 0     \\ 
        \hline
        $u_4$                           & 0       &       & 0     &      & 1      & 1     \\ 
        \hline
    \end{tabularx}
    \caption{Like ratings.}
    \label{tab:example-dataset-split:like}
\end{subtable}
\begin{subtable}{.48\textwidth}
    \centering
    \begin{tabularx}{\textwidth}{ 
      | >{\centering\arraybackslash}X 
      | >{\centering\arraybackslash}X 
      | >{\centering\arraybackslash}X 
      | >{\centering\arraybackslash}X
      | >{\centering\arraybackslash}X 
      | >{\centering\arraybackslash}X 
      | >{\centering\arraybackslash}X | }
        \hline 
        $R_{u,i}^{\text{\faThumbsODown}}$ & $i_1$ & $i_2$ & $i_3$ & $i_4$ & $i_5$ & $i_6$ \\ 
        \hline
        $u_1$                             & 1       & 0     &       & 0    &        &       \\ 
        \hline
        $u_2$                             & 0       &       & 0     & 1    & 1      &       \\ 
        \hline
        $u_3$                             &         & 0     & 0     &      &        & 1     \\ 
        \hline
        $u_4$                             & 1       &       & 1     &      & 0      & 0     \\ 
        \hline
    \end{tabularx}
    \caption{Dislike ratings.}
    \label{tab:example-dataset-split:dislike}
\end{subtable}

\caption{Resulting matrix after splitting the running example's rating matrix.}
\label{tab:example-dataset-split}
\end{table}

During the fitting process, the \ac{BeMF} model must learn its parameters following algorithm~\ref{alg:bemf}. These parameters are stored in four matrices that contain the latent factors of the users for the `like' rating ($U^{\text{\faThumbsOUp}}$), latent factors of the users for the `dislike' rating ($U^{\text{\faThumbsODown}}$), latent factors of the items for the `like' rating ($V^{\text{\faThumbsOUp}}$) and latent factors of the items for the `dislike' rating ($V^{\text{\faThumbsODown}}$). For this example, we have fixed the number of latent factors to $k=3$.~\Cref{tab:example-params-init} contains a random initialization of these parameters for the running example.

\begin{table}[!h]
\centering
\begin{subtable}{.48\textwidth}
    \centering
    \begin{tabularx}{\textwidth}{ 
      | >{\centering\arraybackslash}X 
      | >{\centering\arraybackslash}X 
      | >{\centering\arraybackslash}X 
      | >{\centering\arraybackslash}X
      | >{\centering\arraybackslash}X 
      | >{\centering\arraybackslash}X 
      | >{\centering\arraybackslash}X | }
        \hline 
        $U_{u,f}^{\text{\faThumbsOUp}}$ & $f_1$ & $f_2$ & $f_3$ \\ 
        \hline
        $u_1$                           & 0.99  & 0.26  & 0.55  \\ 
        \hline
        $u_2$                           & 0.77  & 0.77  & 0.85  \\ 
        \hline
        $u_3$                           & 0.20  & 0.27  & 0.35  \\ 
        \hline
        $u_4$                           & 0.11  & 0.96  & 0.13  \\ 
        \hline
    \end{tabularx}
    \caption{Users latent factors for the like rating.}
\end{subtable}
\begin{subtable}{.48\textwidth}
    \centering
    \begin{tabularx}{\textwidth}{ 
      | >{\centering\arraybackslash}X 
      | >{\centering\arraybackslash}X 
      | >{\centering\arraybackslash}X 
      | >{\centering\arraybackslash}X
      | >{\centering\arraybackslash}X 
      | >{\centering\arraybackslash}X 
      | >{\centering\arraybackslash}X | }
        \hline 
        $U_{u,f}^{\text{\faThumbsODown}}$ & $f_1$ & $f_2$ & $f_3$  \\ 
        \hline
        $u_1$                             & 0.61  & 0.83  & 0.47   \\ 
        \hline
        $u_2$                             & 0.12  & 0.02  & 0.54   \\ 
        \hline
        $u_3$                             & 0.11  & 0.41  & 0.07   \\ 
        \hline
        $u_4$                             & 0.81  & 0.92  & 0.52   \\ 
        \hline
    \end{tabularx}
    \caption{Users latent factors for the dislike rating.}
\end{subtable}

\vspace{0.5cm}

\begin{subtable}{.48\textwidth}
    \centering
    \begin{tabularx}{\textwidth}{ 
      | >{\centering\arraybackslash}X 
      | >{\centering\arraybackslash}X 
      | >{\centering\arraybackslash}X 
      | >{\centering\arraybackslash}X
      | >{\centering\arraybackslash}X 
      | >{\centering\arraybackslash}X 
      | >{\centering\arraybackslash}X | }
        \hline 
        $V_{i,f}^{\text{\faThumbsOUp}}$ & $f_1$ & $f_2$ & $f_3$ \\ 
        \hline
        $i_1$                           & 0.91  & 0.15  & 0.27  \\ 
        \hline
        $i_2$                           & 0.54  & 0.54  & 0.79  \\ 
        \hline
        $i_3$                           & 0.31  & 0.57  & 0.09  \\ 
        \hline
        $i_4$                           & 1.00  & 0.83  & 0.75  \\ 
        \hline
        $i_5$                           & 0.68  & 0.03  & 0.05  \\ 
        \hline
        $i_6$                           & 0.35  & 0.50  & 0.75  \\ 
        \hline
    \end{tabularx}
    \caption{Items latent factors for the like rating.}
\end{subtable}
\begin{subtable}{.48\textwidth}
    \centering
    \begin{tabularx}{\textwidth}{ 
      | >{\centering\arraybackslash}X 
      | >{\centering\arraybackslash}X 
      | >{\centering\arraybackslash}X 
      | >{\centering\arraybackslash}X
      | >{\centering\arraybackslash}X 
      | >{\centering\arraybackslash}X 
      | >{\centering\arraybackslash}X | }
        \hline 
        $V_{i,f}^{\text{\faThumbsODown}}$ & $f_1$ & $f_2$ & $f_3$ \\ 
        \hline
        $i_1$                             & 0.92  & 0.53  & 0.67  \\ 
        \hline
        $i_2$                             & 0.40  & 0.24  & 0.12  \\ 
        \hline
        $i_3$                             & 0.64  & 0.22  & 0.89  \\ 
        \hline
        $i_4$                             & 0.64  & 0.86 & 0.60   \\ 
        \hline
        $i_5$                             & 0.51  & 0.12 & 0.41   \\ 
        \hline
        $i_6$                             & 0.92  & 0.23 & 0.75   \\ 
        \hline
    \end{tabularx}
    \caption{Items latent factors for the dislike rating.}
\end{subtable}

\caption{Initial random \ac{BeMF} model parameters in the running example. The number of latent factors has been fixed to $k=3$.}
\label{tab:example-params-init}
\end{table}

To learn the parameters, a gradient descent approach is employed, so we must perform m iterations following the update rules for the latent factors. For example, the update of the first factor $f_1$ of user $u_1$ for the like rating, which is denoted $\Delta U_{u_1, f_1}^{\text{\faThumbsOUp}}$, is computed as

$$
\Delta U_{u_1, f_1}^{\text{\faThumbsOUp}} = \left(1 - logit(U_{u_1} V_{i_2}) \right) V_{i_2,f_1} + \left(1 - logit(U_{u_1} V_{i_4}) \right) V_{i_4,f_1} - logit(U_{u_1} V_{i_1})V_{i_1,f_1}.
$$

From this value, we can improve the hidden factor by updating

$$
U_{u_1, f_1}^{\text{\faThumbsOUp}} \leftarrow U_{u_1, f_1}^{\text{\faThumbsOUp}} + \gamma \left(\Delta U_{u_1, f_1}^{\text{\faThumbsOUp}} - \eta U_{u_1, f_1}^{\text{\faThumbsOUp}} \right).
$$

Note that only the items rated by user $u_1$ (i.e., $i_1$, $i_2$ and $i_4$) update his/her latent factors.

\Cref{tab:example-params-final} contains the latent factors after one iteration ($m=1$) using the learning rate $\gamma=0.1$ and regularization $\eta=0.01$.

\begin{table}[!h]
\centering
\begin{subtable}{.48\textwidth}
    \centering
    \begin{tabularx}{\textwidth}{ 
      | >{\centering\arraybackslash}X 
      | >{\centering\arraybackslash}X 
      | >{\centering\arraybackslash}X 
      | >{\centering\arraybackslash}X
      | >{\centering\arraybackslash}X 
      | >{\centering\arraybackslash}X 
      | >{\centering\arraybackslash}X | }
        \hline 
        $U_{u,f}^{\text{\faThumbsOUp}}$ & $f_1$ & $f_2$ & $f_3$ \\ 
        \hline
        $u_1$                & 0.96  & 0.27  & 0.56  \\ 
        \hline
        $u_2$                & 0.72  & 0.75  & 0.81  \\ 
        \hline
        $u_3$                & 0.21  & 0.29  & 0.34  \\ 
        \hline
        $u_4$                & 0.05  & 0.92  & 0.12  \\ 
        \hline
    \end{tabularx}
    \caption{Users latent factors for the like rating.}
\end{subtable}
\begin{subtable}{.48\textwidth}
    \centering
    \begin{tabularx}{\textwidth}{ 
      | >{\centering\arraybackslash}X 
      | >{\centering\arraybackslash}X 
      | >{\centering\arraybackslash}X 
      | >{\centering\arraybackslash}X
      | >{\centering\arraybackslash}X 
      | >{\centering\arraybackslash}X 
      | >{\centering\arraybackslash}X | }
        \hline 
        $U_{u,f}^{\text{\faThumbsODown}}$ & $f_1$ & $f_2$ & $f_3$  \\ 
        \hline
        $u_1$                             & 0.55  & 0.76  & 0.43   \\ 
        \hline
        $u_2$                             & 0.07  & 0.02  & 0.49   \\ 
        \hline
        $u_3$                             & 0.10  & 0.39  & 0.05   \\ 
        \hline
        $u_4$                             & 0.73  & 0.90  & 0.46   \\ 
        \hline
    \end{tabularx}
    \caption{Users latent factors for the dislike rating.}
\end{subtable}

\vspace{0.5cm}

\begin{subtable}{.48\textwidth}
    \centering
    \begin{tabularx}{\textwidth}{ 
      | >{\centering\arraybackslash}X 
      | >{\centering\arraybackslash}X 
      | >{\centering\arraybackslash}X 
      | >{\centering\arraybackslash}X
      | >{\centering\arraybackslash}X 
      | >{\centering\arraybackslash}X 
      | >{\centering\arraybackslash}X | }
        \hline 
        $V_{i,f}^{\text{\faThumbsOUp}}$ & $f_1$ & $f_2$ & $f_3$ \\ 
        \hline
        $i_1$                           & 0.82  & 0.13  & 0.24  \\ 
        \hline
        $i_2$                           & 0.58  & 0.58  & 0.84  \\ 
        \hline
        $i_3$                           & 0.31  & 0.58  & 0.09  \\ 
        \hline
        $i_4$                           & 0.95  & 0.79  & 0.71  \\ 
        \hline
        $i_5$                           & 0.66  & 0.03  & 0.05  \\ 
        \hline
        $i_6$                           & 0.34  & 0.48  & 0.72  \\ 
        \hline
    \end{tabularx}
    \caption{Items latent factors for the like rating.}
\end{subtable}
\begin{subtable}{.48\textwidth}
    \centering
    \begin{tabularx}{\textwidth}{ 
      | >{\centering\arraybackslash}X 
      | >{\centering\arraybackslash}X 
      | >{\centering\arraybackslash}X 
      | >{\centering\arraybackslash}X
      | >{\centering\arraybackslash}X 
      | >{\centering\arraybackslash}X 
      | >{\centering\arraybackslash}X | }
        \hline 
        $V_{i,f}^{\text{\faThumbsODown}}$ & $f_1$ & $f_2$ & $f_3$ \\ 
        \hline
        $i_1$                             & 0.90  & 0.52  & 0.66  \\ 
        \hline
        $i_2$                             & 0.36  & 0.21  & 0.11  \\ 
        \hline
        $i_3$                             & 0.57  & 0.20  & 0.80  \\ 
        \hline
        $i_4$                             & 0.61  & 0.82  & 0.58   \\ 
        \hline
        $i_5$                             & 0.50  & 0.12  & 0.40   \\ 
        \hline
        $i_6$                             & 0.89  & 0.22  & 0.73   \\ 
        \hline
    \end{tabularx}
    \caption{Items latent factors for the dislike rating.}
\end{subtable}

\caption{\ac{BeMF} model parameters after one iteration in the running example.}
\label{tab:example-params-final}
\end{table}

Once the \ac{BeMF} model has been trained, predictions can be computed by obtaining the score that maximizes the probability in the classification task. For example, to predict the rating of user $u_1$ to the item $i_3$, $\hat{R}_{u_1, i_3}$, we must compute the probability distribution of this rating, $\Phi(u_1, i_3)$, as

\begin{align}
\Phi(u_1, i_3) &= \frac{1}{logit \left( U_{u_1}^{\text{\faThumbsODown}} V_{i_3}^{\text{\faThumbsODown}} \right) + logit \left( U_{u_1}^{\text{\faThumbsOUp}} V_{i_3}^{\text{\faThumbsOUp}} \right) }  \left( logit \left( U_{u_1}^{\text{\faThumbsODown}} V_{i_3}^{\text{\faThumbsODown}} \right) , logit \left( U_{u_1}^{\text{\faThumbsOUp}} V_{i_3}^{\text{\faThumbsOUp}} \right) \right) \\
&= \frac{1}{0.71+0.62} \left( 0.71, 0.62 \right) = \left( 0.53, 0.47 \right).
\end{align}

Henceforth, the \ac{BeMF} model will return $\hat{R}_{u_1, i_3}=\text{\textcolor{Red}{\faThumbsODown}}$ with the reliability $\rho_{u_1,i_3}=0.53$.

\section{Proposed model evaluation}\label{sec:evaluation}

This section contains a detailed explanation of the experiments that are carried out to evaluate the proposed model. \Cref{sec:evaluation:setup} describes the experimental setup that defines the datasets, baselines and quality measures that are employed during the evaluation. \Cref{sec:evaluation:results} includes the experimental results and a comparison of the performance of the proposed method regarding the selected baselines. All experiments have been conducted using \ac{CF4J}~\cite{ortega2018cf4j}; their source code is available at \url{https://github.com/ferortega/bernoulli-matrix-factorization}.

\subsection{Experimental setup}\label{sec:evaluation:setup}

Experimental evaluation was conducted using the MovieLens~\cite{harper2015movielens}, FilmTrust~\cite{guo2013novel} and MyAnimeList datasets. These datasets have been selected to assess the impact of splitting the rating matrix into binary rating matrices using different discrete sets of possible scores. In this way, the MovieLens dataset contains ratings from 1 star to 5 stars FilmTrust ratings range from 0.5 to 4.0 with half increments; and the MyAnimeList dataset restricts its possible scores to the range 1 to 10. Moreover, to ensure the reproducibility of these experiments, all of them were carried out using the benchmark version of these datasets, which are included in \ac{CF4J}~\cite{ortega2018cf4j}. The main features of these datasets are shown in \cref{tab:datasets}.

\begin{table}[h]
\begin{tabularx}{\textwidth}{ 
  | >{\raggedright\arraybackslash}X 
  | >{\centering\arraybackslash}X 
  | >{\centering\arraybackslash}X
  | >{\centering\arraybackslash}X 
  | >{\centering\arraybackslash}X 
  | >{\centering\arraybackslash}X | }
    \hline
    Dataset     & Number of users & Number of items & Number of ratings & Number of test ratings & Possible scores                 \\ 
    \hline
    \small MovieLens   & 6,040           & 3,706           & 911,031           & 89,178                 & 1 to 5 stars                    \\ 
    \hline
    \small FilmTrust   & 1,508           & 2,071           & 32,675            & 2,819                  & 0.5 to 4.0 with half increments \\ 
    \hline
    \small MyAnimeList & 69,600          & 9,927           & 5,788,207         & 549,027                & 1 to 10                         \\ 
    \hline
\end{tabularx}
\caption{Main parameters of the datasets used in the experiments.}
\label{tab:datasets}
\end{table}

According to \cref{sec:introduction}, all the \ac{MF} models that are presented in the literature are capable of estimating a user’s rating prediction for an item, but only a few \ac{MF} models provide the reliability of their predictions and recommendations. Baselines have been selected to supply a heterogeneous representation of all the existing \ac{MF} models. \Cref{tab:baselines-outputs} contains the selected baselines and their generated outputs. Note that the proposed model, \ac{BeMF}, can estimate the user’s rating predictions and compute the reliability of the predictions and recommendations, as described in \cref{sec:bemf:model}.

\begin{table}[h]
\begin{tabularx}{\textwidth}{ 
  | >{\raggedright\arraybackslash}X 
  | >{\centering\arraybackslash}X 
  | >{\centering\arraybackslash}X 
  | >{\centering\arraybackslash}X | }
    \hline
    Model                            & Prediction                & Prediction reliability    & Recommendation reliability \\ 
    \hline
    BiasedMF~\cite{Koren2009Aug}     & \textcolor{Green}{\cmark} & \textcolor{Red}{\xmark}   & \textcolor{Red}{\xmark}    \\ 
    \hline
    BNMF~\cite{Hernando2016Apr}      & \textcolor{Green}{\cmark} & \textcolor{Red}{\xmark}   & \textcolor{Green}{\cmark}  \\
    \hline
    NMF~\cite{NIPS2000_1861}         & \textcolor{Green}{\cmark} & \textcolor{Red}{\xmark}   & \textcolor{Red}{\xmark}    \\ 
    \hline
    PMF~\cite{mnih2008probabilistic} & \textcolor{Green}{\cmark} & \textcolor{Red}{\xmark}   & \textcolor{Red}{\xmark}    \\
    \hline
    SVD++\cite{Koren2008}            & \textcolor{Green}{\cmark} & \textcolor{Red}{\xmark}   & \textcolor{Red}{\xmark}    \\
    \hline
    URP~\cite{Marlin2003}            & \textcolor{Green}{\cmark} & \textcolor{Green}{\cmark} & \textcolor{Green}{\cmark}  \\ 
    \hline
\end{tabularx}
\caption{Output generated by \ac{MF} models selected as baselines.}
\label{tab:baselines-outputs}
\end{table}

The chosen baselines contain several hyperparameters that must be tuned. We analyzed the prediction error of the baseline models with different values of these hyperparameters by performing a grid search optimization that minimizes the mean absolute prediction error. \Cref{tab:baselines-hyperparams} contains the hyperparameters that are generated from this optimization process for each baseline and dataset.

\begin{table}[h]
\begin{tabularx}{\textwidth}{ 
  | >{\raggedright\arraybackslash}X 
  | >{\raggedright\arraybackslash}X 
  | >{\raggedright\arraybackslash}X 
  | >{\raggedright\arraybackslash}X | }
    \hline
    Method   & MovieLens                                             & FilmTrust                                             & MyAnimeList                                            \\ 
    \hline
    PMF      & $\textrm{factors}=8$, $\gamma=0.01$, $\lambda=0.045$  & $\textrm{factors}=4$, $\gamma=0.015$, $\lambda=0.1$   & $\textrm{factors}=10$, $\gamma=0.005$, $\lambda=0.085$ \\ 
    \hline
    BiasedMF & $\textrm{factors}=6$, $\gamma=0.01$, $\lambda=0.055$  & $\textrm{factors}=2$, $\gamma=0.015$, $\lambda=0.15$  & $\textrm{factors}=10$, $\gamma=0.01$, $\lambda=0.085$  \\ 
    \hline
    NMF      & $\textrm{factors}=2$                                  & $\textrm{factors}=2$                                  & $\textrm{factors}=2$                                   \\ 
    \hline
    BNMF     & $\textrm{factors}=10$, $\alpha=0.6$, $\beta=5$        & $\textrm{factors}=10$, $\alpha=0.4$, $\beta=25$       & $\textrm{factors}=4$, $\alpha=0.5$, $\beta=5$          \\ 
    \hline
    URP      & $\textrm{factors}=10$                                 & $\textrm{factors}=4$                                  & $\textrm{factors}=8$                                   \\ 
    \hline
    SVD++    & $\textrm{factors}=4$, $\gamma=0.0014$, $\lambda=0.05$ & $\textrm{factors}=2$, $\gamma=0.0014$, $\lambda=0.02$ & $\textrm{factors}=4$, $\gamma=0.0015$, $\lambda=0.1$   \\ \hline
\end{tabularx}
\caption{Baseline hyperparameters generated from grid search optimization.} 
\label{tab:baselines-hyperparams}
\end{table}

In the same way, the \ac{BeMF} model has some hyperparameters that must be tuned: $k$, which denotes the number of latent factors of the model; $\gamma$, which represents the learning rate of the gradient descent optimization algorithm; $\eta$, which controls the regularization to avoid overfitting; and $m$, the number of iterations of the fitting process. According to algorithm~\ref{alg:bemf}, we have fixed the activation function to the logistic function. As with the baselines, a grid search optimization was carried out to compare the prediction error of the proposed model using thousands of combinations of the required hyperparameters. We evaluated the following intervals of values for each hyperparameter:

\begin{itemize}
    \item $k$: from 2 to 8 latent factors with increments of 2.
    \item $\gamma$: from 0.002 to 0.02 with increments of 0.002.
    \item $\eta$: from 0.01 to 0.2 with increments of 0.01.
    \item $m$: from 50 to 100 with increments of 25.
\end{itemize}

The total number of all the possible combinations of these parameter values is 2400. In the MovieLens dataset, the minimum prediction error was obtained using $k=2$, $\gamma=0.006$, $\lambda=0.16$ and $m=100$ iterations. In the same way, in the FilmTrust dataset, the hyperparameter values that minimize the prediction error are $k=2$, $\gamma=0.02$, $\lambda=0.06$ and $m=75$ iterations. In MyAnimeList, the best predictions are estimated using $k=4$, $\gamma=0.004$, $\lambda=0.1$ and $m=100$ iterations.

The quality of the predictions and recommendations provided by a \ac{CF}-based \ac{RS} must be evaluated using standard quality measures. To measure the quality of the predictions, we define \ac{MAE} (\cref{eq:mae}) as the mean absolute difference between the test ratings ($R_{u,i}$) and their predictions ($\hat{R}_{u,i}$) and define coverage (\cref{eq:coverage}) as the proportion of test ratings that a \ac{CF} can predict ($\hat{R}_{u,i} \neq \bullet$) concerning the total number of test ratings. Here, $R^{test}$ is the collection of pairs $\langle u,i\rangle$ of user $u$ and item $i$ in the test split of the dataset, and $\#R^{test}$ denotes its cardinality.

\begin{equation} \label{eq:mae}
    \textrm{MAE} = \frac{1}{\#R^{test}} \sum_{\langle u,i \rangle \in R^{test}} \lvert R_{u,i} - \hat{R}_{u,i} \rvert.
\end{equation}

\begin{equation} \label{eq:coverage}
    \textrm{Coverage} = \frac{ \#\{ \langle u,i \rangle \in R^{text} | \hat{R}_{u,i} \neq \bullet \}}{\#R^{test}}.
\end{equation}

On the other hand, fixing $n > 0$ to measure the quality of the top $n$ recommendations, we can consider two adapted quality measures. The first quality measure is precision, which is given by the averaged proportion of successful recommendations that are included in the recommendation list of user $u$ of maximum length $n$, which is denoted $T^n_u$, with respect to the size of the recommendation list (\cref{eq:precision}). The second quality measure is recall, which is the averaged proportion of successful recommendations included in the recommendation list of user $u$, $T^n_u$, with respect to the total number of test items that user $u$ likes (\cref{eq:recall}).

\begin{equation} \label{eq:precision}
    \textrm{precision} = \frac{1}{N} \sum_{u = 1}^N \frac{\{ i \in T^n_u | R_{u,i} \geq \theta \}}{\#T^n_u}.
\end{equation}

\begin{equation} \label{eq:recall}
    \textrm{recall} = \frac{1}{N} \sum_{u=1}^N \frac{\{ i \in T^n_u | R_{u,i} \geq \theta \}}{\{ i \in R^{test}_u | R_{u,i} \geq \theta \}}.
\end{equation}

In the previous formulae, $u$ runs over the users of the dataset, $N$ is the total number of users, $R^{test}_u$ is the collection of items rated by user $u$ in the test split and $\theta$ is a threshold to discern if a user likes an item ($R_{u,i} \geq \theta$) or not ($R_{u,i} < \theta$).

\subsection{Experimental results}\label{sec:evaluation:results}

\begin{figure}[h]
\centering%
\subcaptionbox{MovieLens}{\includegraphics[width=.33\textwidth]{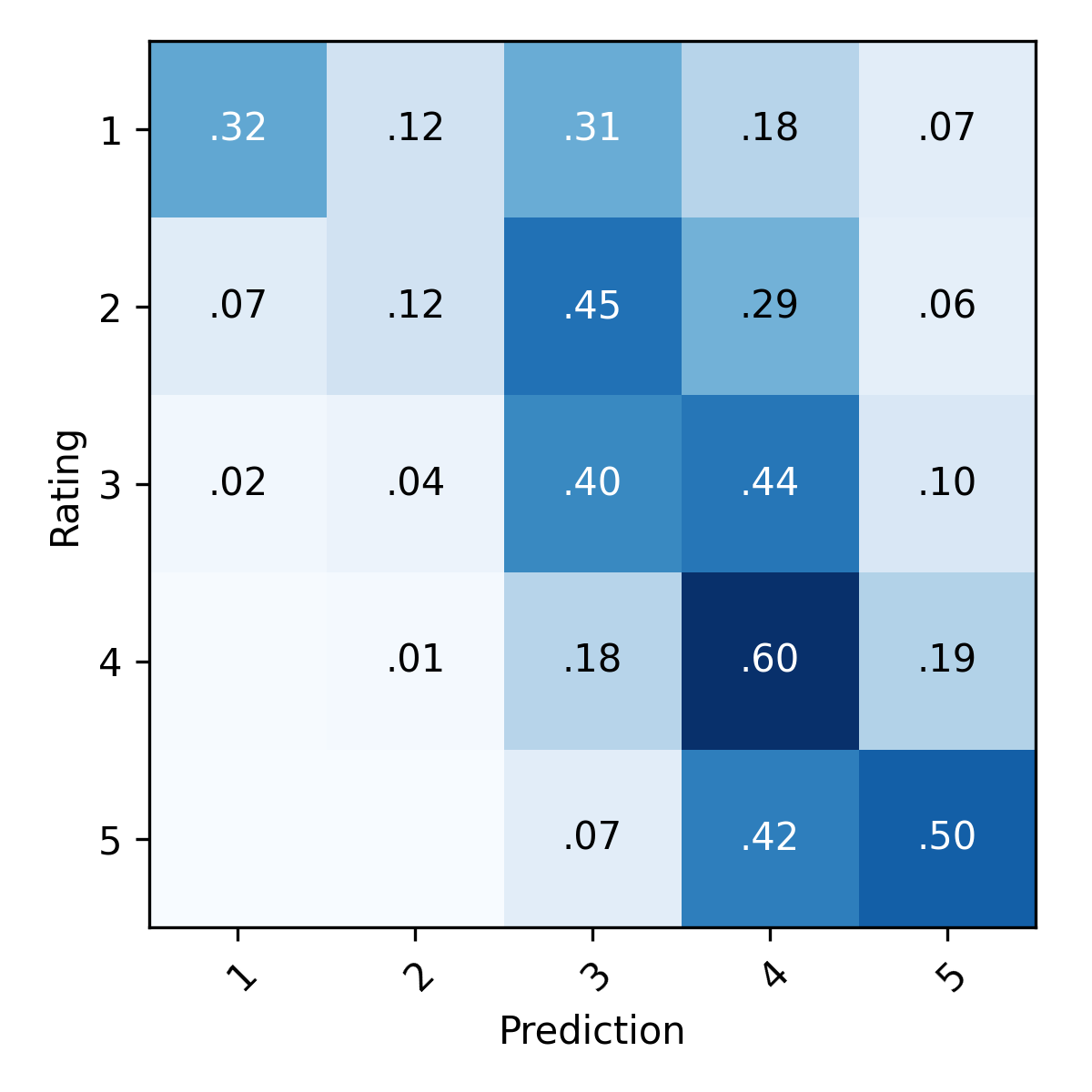}}\hspace{.1em}%
\subcaptionbox{FilmTrust}{\includegraphics[width=.33\textwidth]{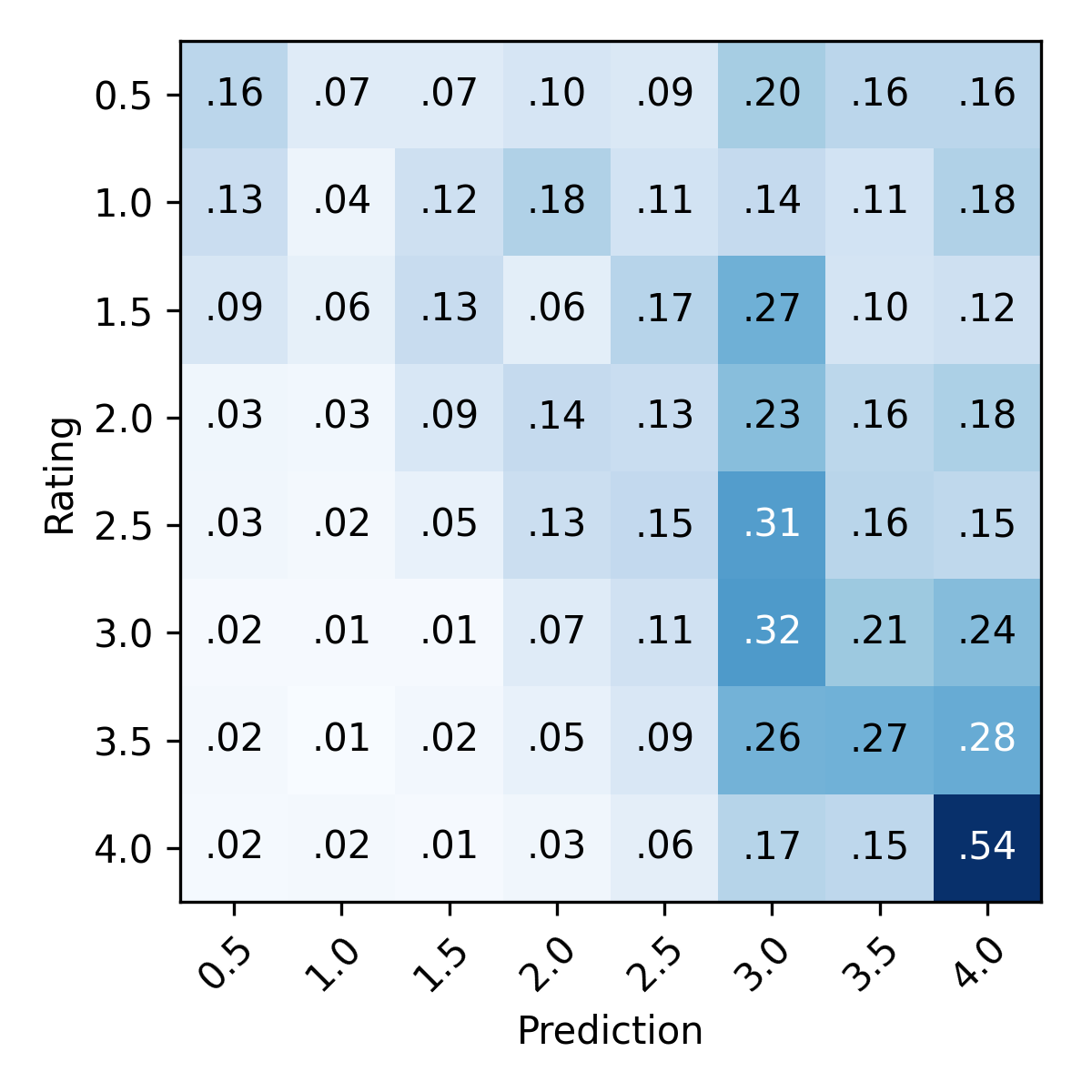}}\hspace{.1em}%
\subcaptionbox{MyAnimeList}{\includegraphics[width=.33\textwidth]{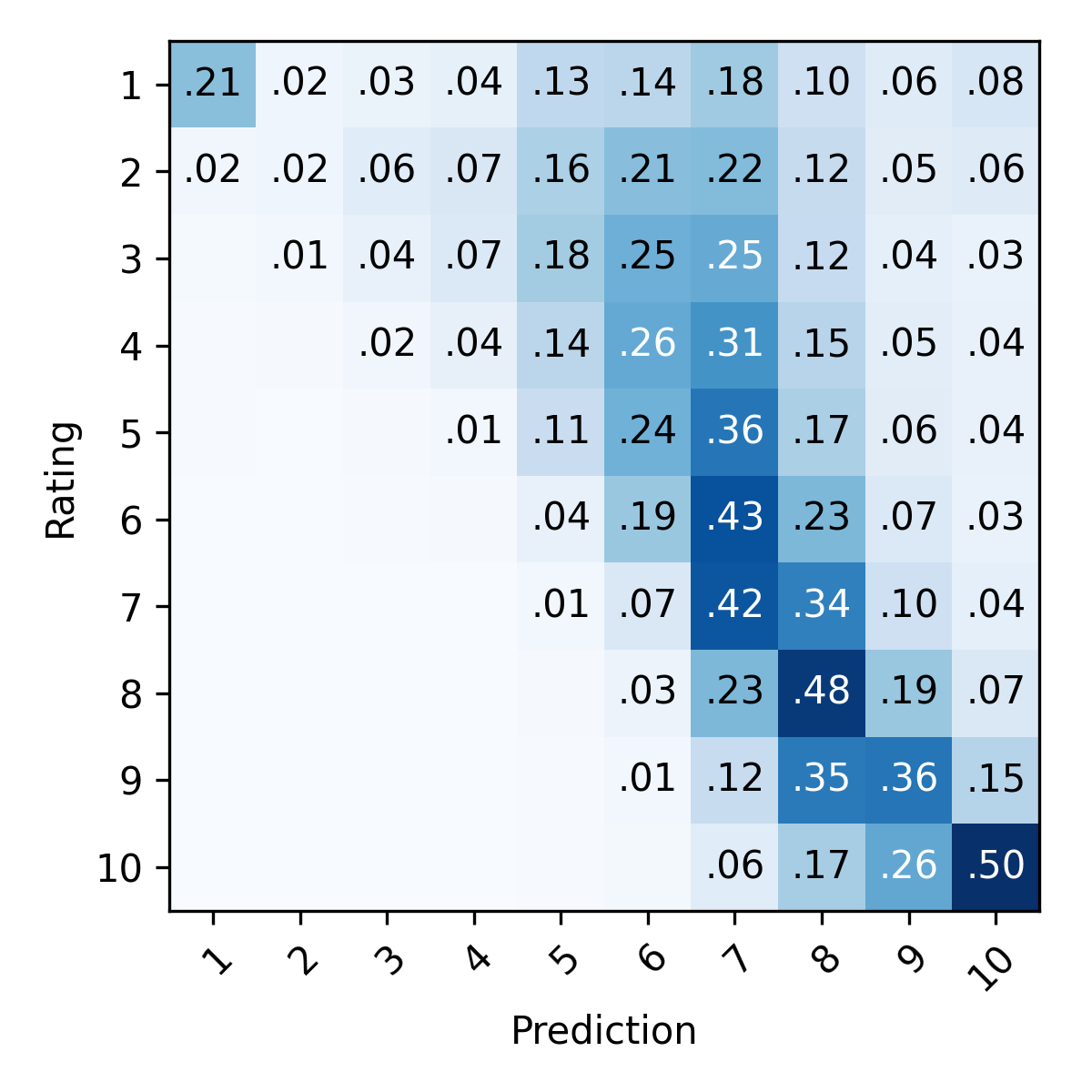}}%
\caption{Confusion matrix of the classification performed by \ac{BeMF}.}
\label{fig:conf-matrix}
\end{figure}

The most popular \ac{MF} based \ac{CF} methods are constructed as regressors because they assume that rating values are continuous, and consequently, their predictions are real values. Conversely, the \ac{BeMF} model has been designed to work with discrete rating values so that it solves the classification problem in which the classes are the possible scores with which items are rated. A confusion matrix is a graphical representation tool that is used to evaluate the output of a classifier. Columns contain the predicted labels, rows contain the real labels and cells denote the proportion of samples of a label that have been predicted with another label. \Cref{fig:conf-matrix} contains the confusion matrix of the \ac{BeMF} model’s predictions for the MovieLens (a), FilmTrust (b) and MyAnimeList (c) datasets. We observe that most of the predictions are correct. However, the model tends to return predictions for high scores due to the bias of the datasets employed in \ac{CF}, where users tend to rate only items that they like.

\begin{figure}[h]
      \centering
      \subcaptionbox{MovieLens}{\includegraphics[width=.33\textwidth]{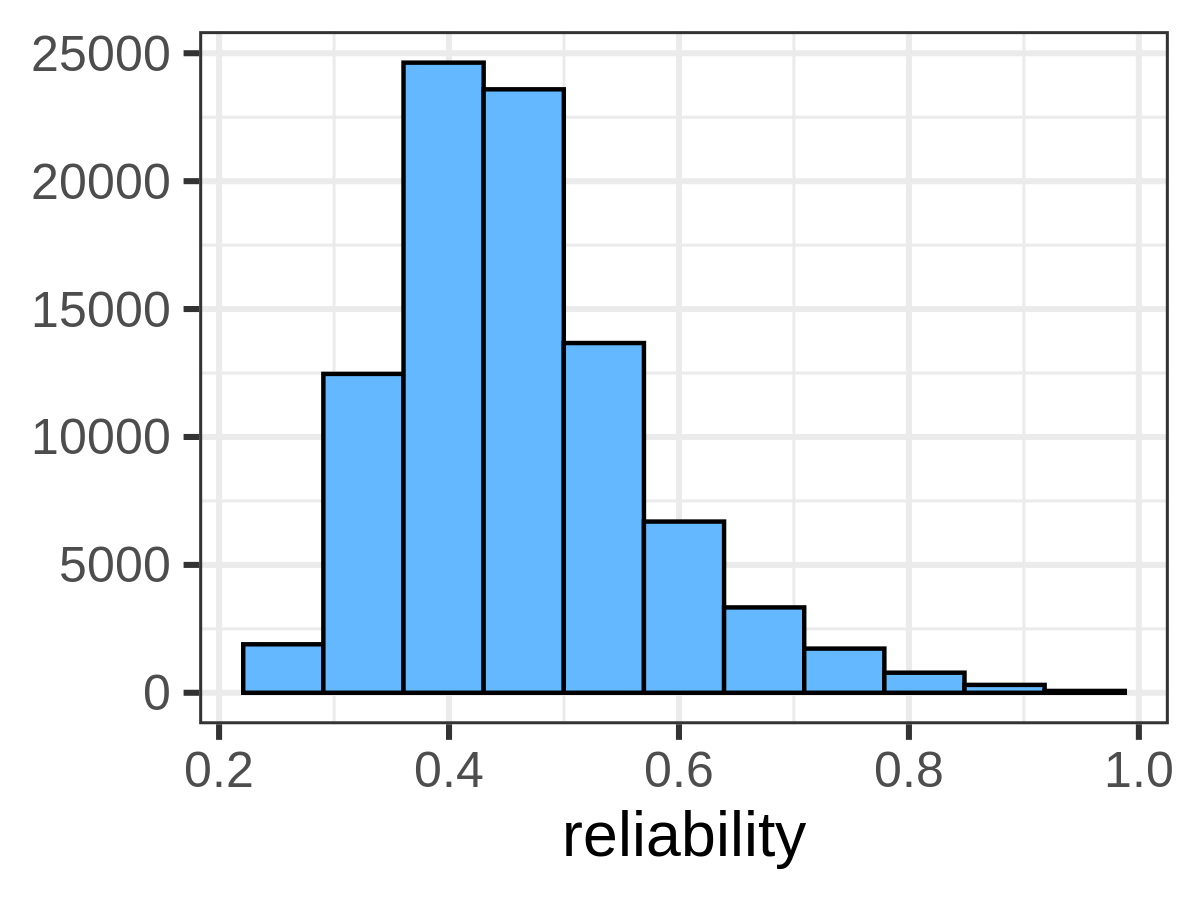}}\hspace{.1em}%
      \subcaptionbox{FilmTrust}{\includegraphics[width=.33\textwidth]{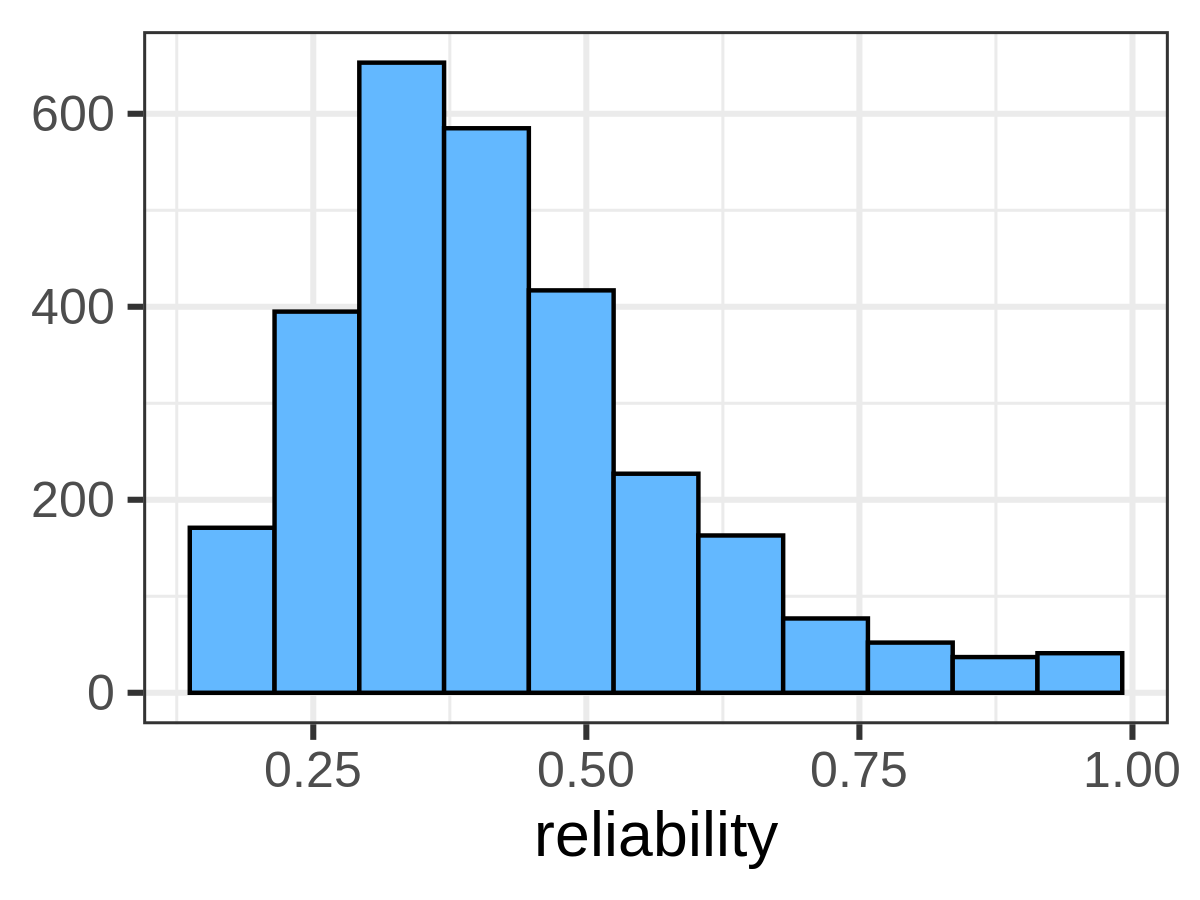}}\hspace{.1em}%
      \subcaptionbox{MyAnimeList}{\includegraphics[width=.33\textwidth]{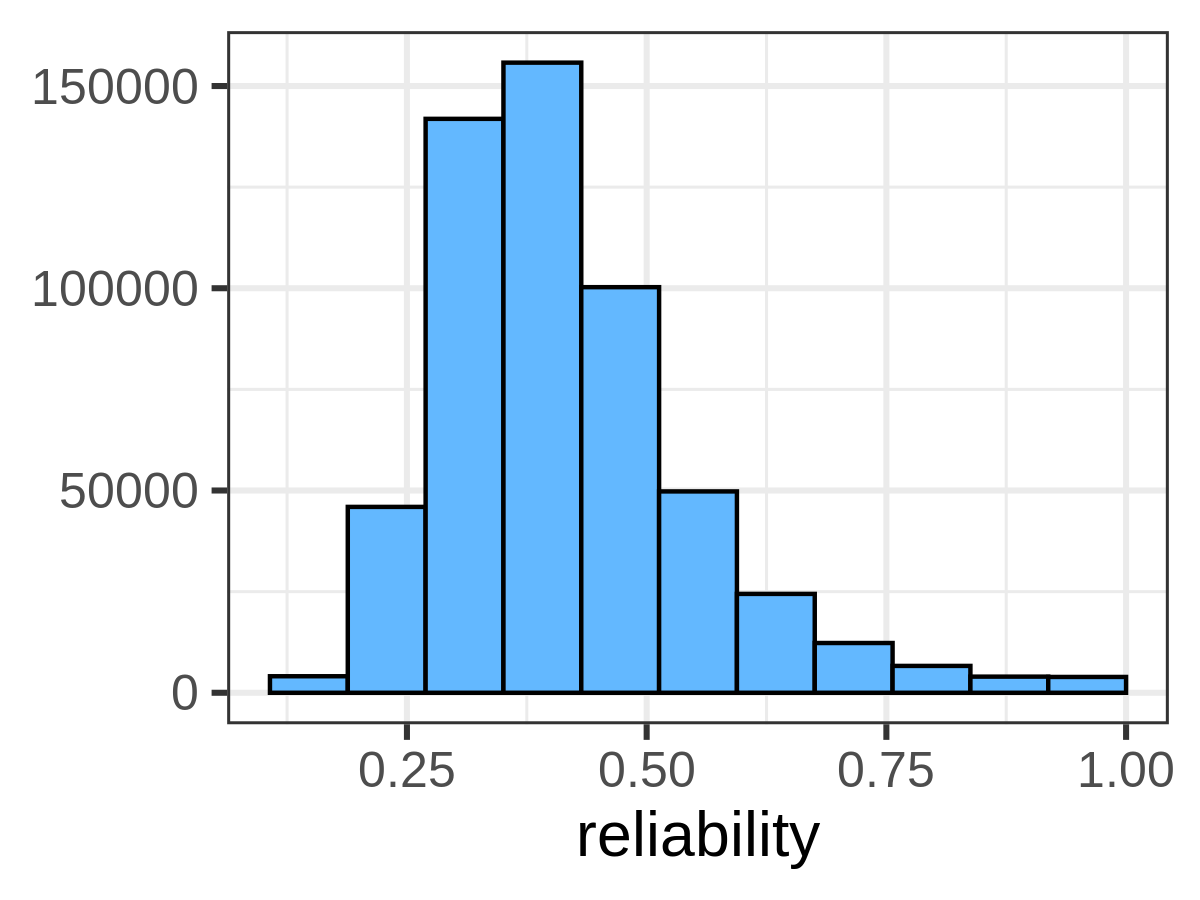}}
    \caption{Histogram of the reliability values of \ac{BeMF}'s test predictions.}
    \label{fig:reliability-histogram}
\end{figure}

The \ac{BeMF} model is capable of providing not only rating prediction but also the reliability of this prediction. The reliability values indicate the model’s confidence in the predictions. \Cref{fig:reliability-histogram} shows the histograms with the distribution of the reliability values of the test predictions in the MovieLens (a), FilmTrust (b) and MyAnimeList (c) datasets. These histograms show that the reliability distribution shape is the same for the three tested datasets: most of the reliability values belong to the interval $[0.3, 0.5]$, which means a confidence interval between $30\%$ to $50\%$ in the predicted rating, and there are few reliabilities with values greater than $0.75$.

The reliability value allows us to calibrate the output of the \ac{BeMF} model. By filtering out less reliable predictions, we decrease the coverage of the model (i.e., some predictions cannot be issued because the model does not have enough confidence to make them) but increase the prediction accuracy. It is reasonable to think that predictions with high reliability are more obvious than predictions with low reliability. For example, if a user has positively rated \textit{Star Wars Episode IV (A New Hope)} and \textit{Star Wars Episode V (The Empire Strikes Back)}, the model will have high confidence in the positive interest of the user to \textit{Star Wars Episode VI (Return of the Jedi)} and will assign a high reliability value to this prediction. Conversely, the same model will have less confidence in the interest of the user in other sci-fi movies, such as \textit{Interstellar} or \textit{Gravity}, and will assign lower reliability values to these predictions.

\begin{figure}[ht]
    \includegraphics[width=\textwidth]{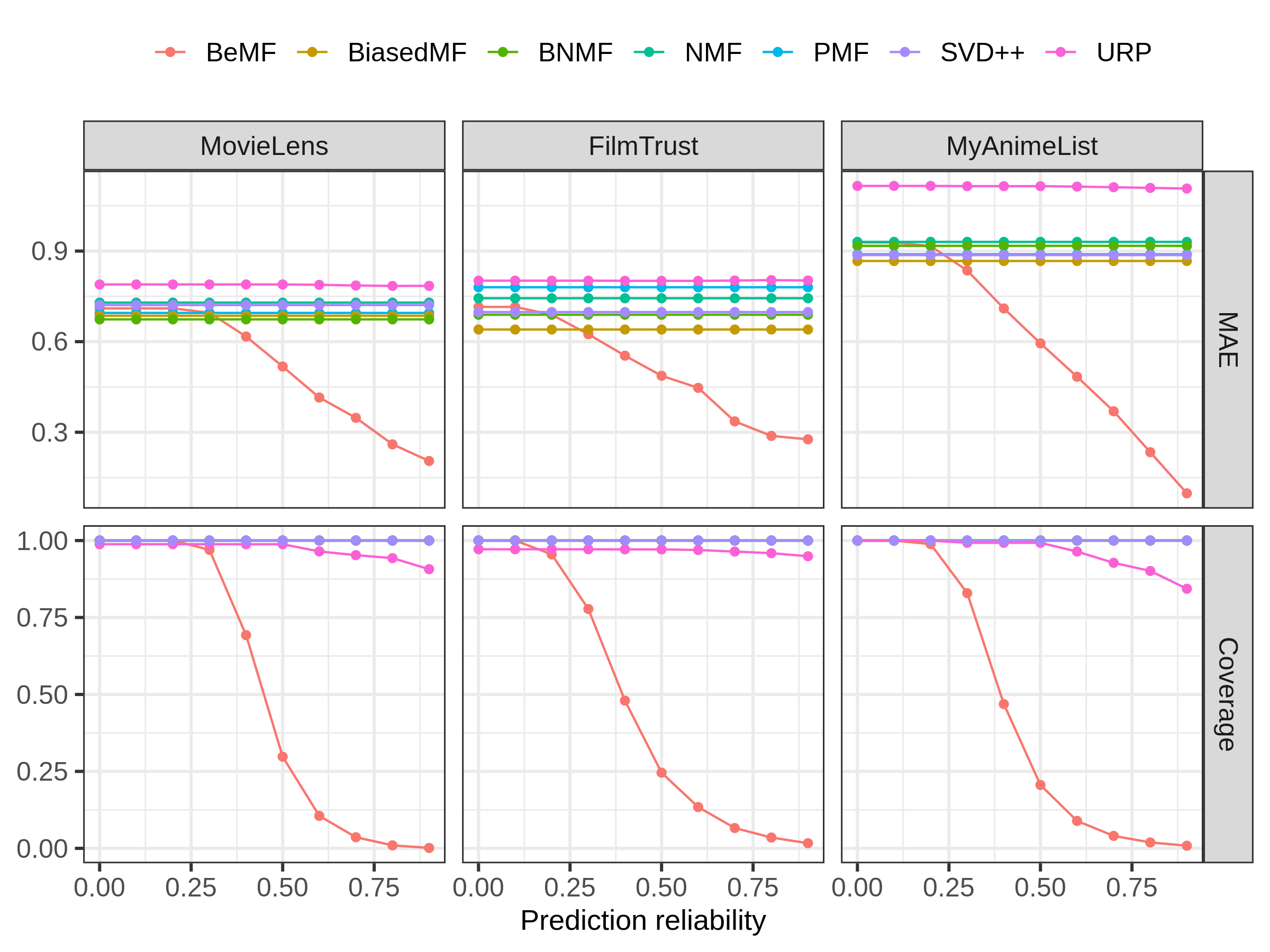}
    \caption{Quality of the predictions measured by \ac{MAE} and \emph{coverage}. Predictions with lower reliability than those indicated on the $x$-axis are filtered out.}
    \label{fig:mae-vs-coverage}
\end{figure}

\Cref{fig:mae-vs-coverage} analyzes the impact of the reliability measure on the quality of the predictions. We have contrasted the prediction error against the predictability of the model using the \ac{MAE} and coverage quality measures that are defined in \cref{eq:mae,eq:coverage}, respectively. The plots were generated by filtering all the predictions with lower reliability than the reliability denoted on the x-axis. Note that only the models that return prediction reliabilities (refer to \cref{tab:baselines-outputs}) can filter their predictions; the remaining predictions are shown as a horizontal line in the plot. All the plots exhibit the same trend: The prediction error is reduced when unreliable predictions are filtered out. Consequently, the coverage of the model also decreases. In any case, the prediction accuracy improvement of the \ac{BeMF} model with respect to the evaluated baselines is significant when the coverage is between $50\%$ and $75\%$.

Despite the satisfactory results reported by the \ac{BeMF} model in the previous experiment, we consider it unfair because not some recommendation models do not benefit from filtering out unreliable predictions. Several methods exist to extend \ac{CF}-based \acp{RS}, which add reliability to their predictions. One of the most popular methods is presented in~\cite{zhu2018assigning}. In this paper, the authors propose computing an auxiliary \ac{MF} model that factorizes a matrix that contains the prediction errors. The authors claim that high reliability values are produced by low prediction errors, and vice-versa.

\begin{figure}[ht]
    \includegraphics[width=\textwidth]{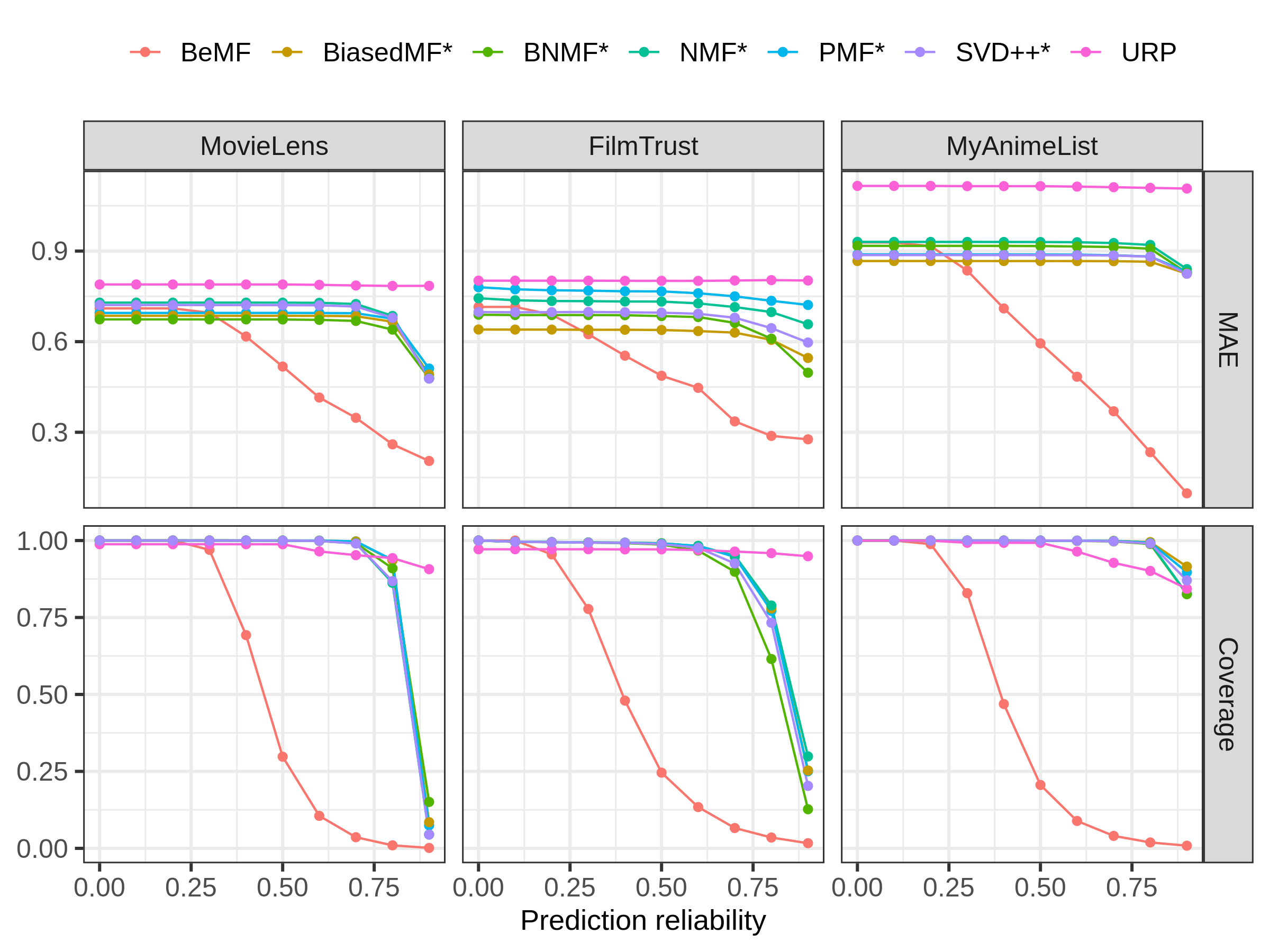}
    \caption{Quality of the predictions measured by \ac{MAE} and \emph{coverage}. Predictions with lower reliability than those indicated on the $x$-axis are filtered out. Models with the suffix * obtain their reliability values using~\cite{zhu2018assigning}.}
    \label{fig:mae-vs-coverage-reliabilized}
\end{figure}

\Cref{fig:mae-vs-coverage-reliabilized} contains the results of repeating the experiment shown in \cref{fig:mae-vs-coverage} but adding a reliability value to the predictions performed by BiasedMF, BNMF, NMF, PMF and SVD++ using~\cite{zhu2018assigning}. The trend observed in the previous experiment is again observed: When unreliable predictions are filtered out, the error decreases and the coverage decreases. However, the reliability values enforced by~\cite{zhu2018assigning} do not achieve results that are as satisfactory as those achieved by the native reliability implemented by \ac{BeMF}.

At this point, it is reasonable to consider that the tuple $\langle \textrm{prediction}, \textrm{reliability}\rangle$ that is returned by the \ac{BeMF} model substantially improves the quality of predictions thanks to the accuracy of the reliability values. To confirm the accuracy of the native reliability values returned by the \ac{BeMF} model, we have compared them against the reliability values enforced by~\cite{zhu2018assigning} when applied to the \ac{BeMF} model’s predictions. To compare both quality measures, we will use \ac{RPI}~\cite{bobadilla2018reliability}. \Cref{tab:rpi} contains the results of this experiment. We can observe that the native \ac{BeMF} reliability values significantly improve those enforced by~\cite{zhu2018assigning}.

\begin{table}[h]
\begin{tabularx}{\textwidth}{ 
  | >{\raggedright\arraybackslash}X 
  | >{\centering\arraybackslash}X 
  | >{\centering\arraybackslash}X | }
    \hline
    Dataset     & \ac{BeMF} (native) & \ac{BeMF} (enforced) \\ 
    \hline
    MovieLens   & 0.09344191         & 0.03607168           \\ 
    \hline
    FilmTrust   & 0.17187947         & 0.03277490           \\ 
    \hline
    MyAnimeList & 0.17087788         & 0.03386983           \\ 
    \hline
\end{tabularx}
\caption{\ac{RPI} values for the native reliability values provided by \ac{BeMF} and the reliability values enforced by~\cite{zhu2018assigning}.} 
\label{tab:rpi}
\end{table}

\begin{figure}[ht]
    \includegraphics[width=\textwidth]{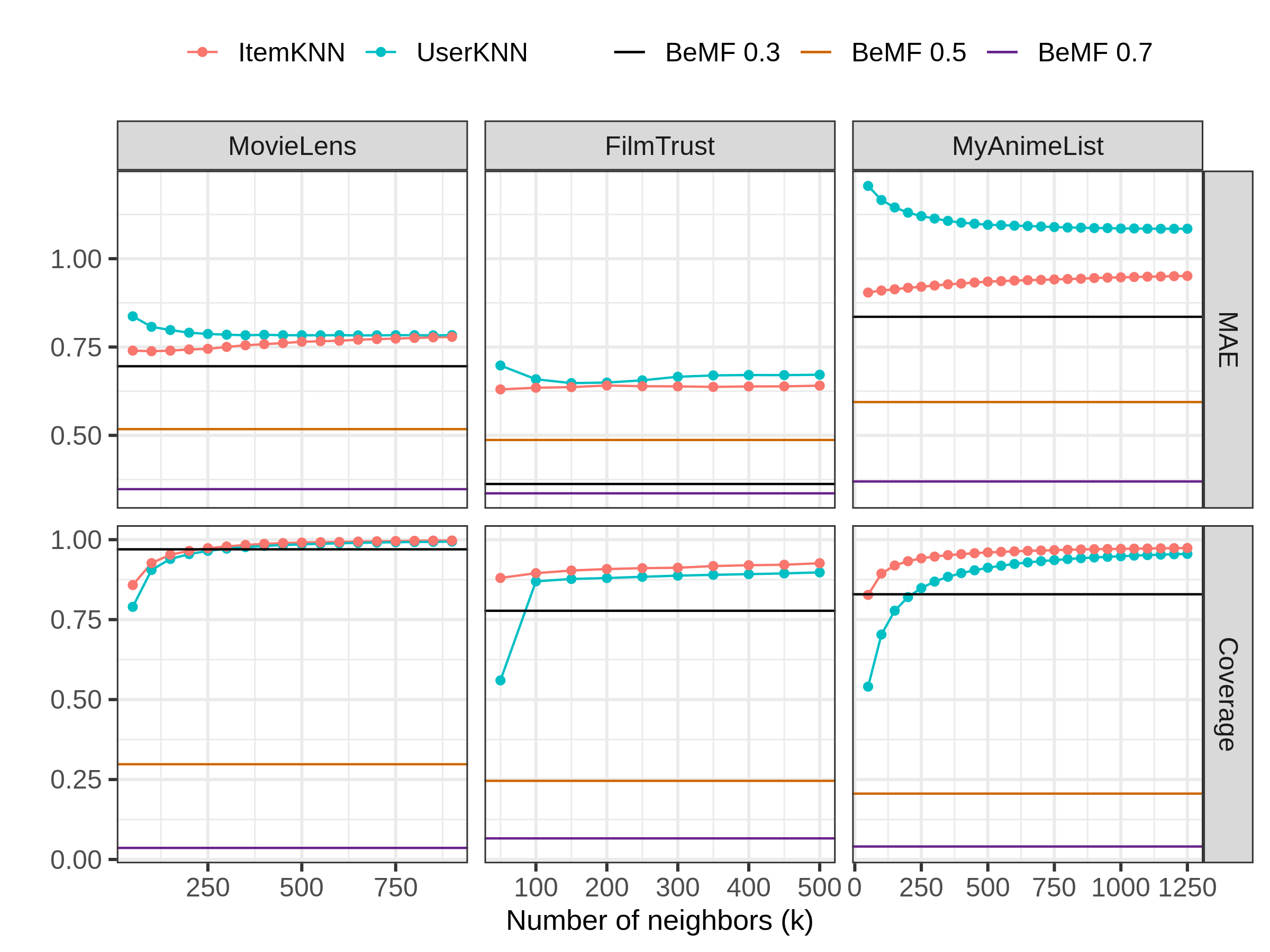}
    \caption{Comparison of the quality of the predictions returned by \ac{CF} based on \ac{KNN} with respect to \ac{BeMF}. \ac{BeMF} is shown by filtering out predictions with a reliability lower than 0.3 (BeMF 0.3), 0.5 (BeMF 0.5) or 0.7 (BeMF 0.7).}
    \label{fig:bemf-vs-knn}
\end{figure}

Previous experiments demonstrate that the satisfactory results are a consequence of including the reliability of the predictions in the model output. As we previously stated, \ac{BeMF} can be calibrated to return few but very reliable predictions or many but less reliable predictions. This property is very innovative in \ac{MF} but not in \ac{CF}. \ac{KNN}-based \ac{CF} systems contain a hyperparameter that tunes the number of neighbors that are considered to issue the prediction. If a low number of neighbors is selected, few predictions are returned, but they are more accurate. Conversely, if a high number of neighbors is selected, many predictions are returned, but they are less accurate. \ac{BeMF} should significantly improve the quality of the predictions returned by \ac{KNN}. To evaluate this hypothesis, \ac{BeMF} has been compared with user-based \ac{KNN} and item-based \ac{KNN} using JMSD~\cite{bobadilla2010new} as a similarity metric. The results of this comparison are shown in \cref{fig:bemf-vs-knn}. We observe that the predictions returned by BeMF have less error than the predictions returned by KNN-based methods, independent of whether the predictions returned by \ac{BeMF} have been filtered with a reliability lower than 0.3 (BeMF 0.3), 0.5 (BeMF 0.5) or 0.7 (BeMF 0.7).

To analyze the quality of the recommendations, we measured the precision (\cref{eq:precision}) and recall (\cref{eq:recall}) of the top 10 recommendations ($n=10$). To discern whether a recommendation is right or wrong, we have fixed the threshold that determines the items that interest a user based on his/her test ratings to $\theta=4$ in MovieLens, $\theta=3.5$ in FilmTrust and $\theta=7$ in MyAnimeList.

\begin{figure}[!h]
    \includegraphics[width=\textwidth]{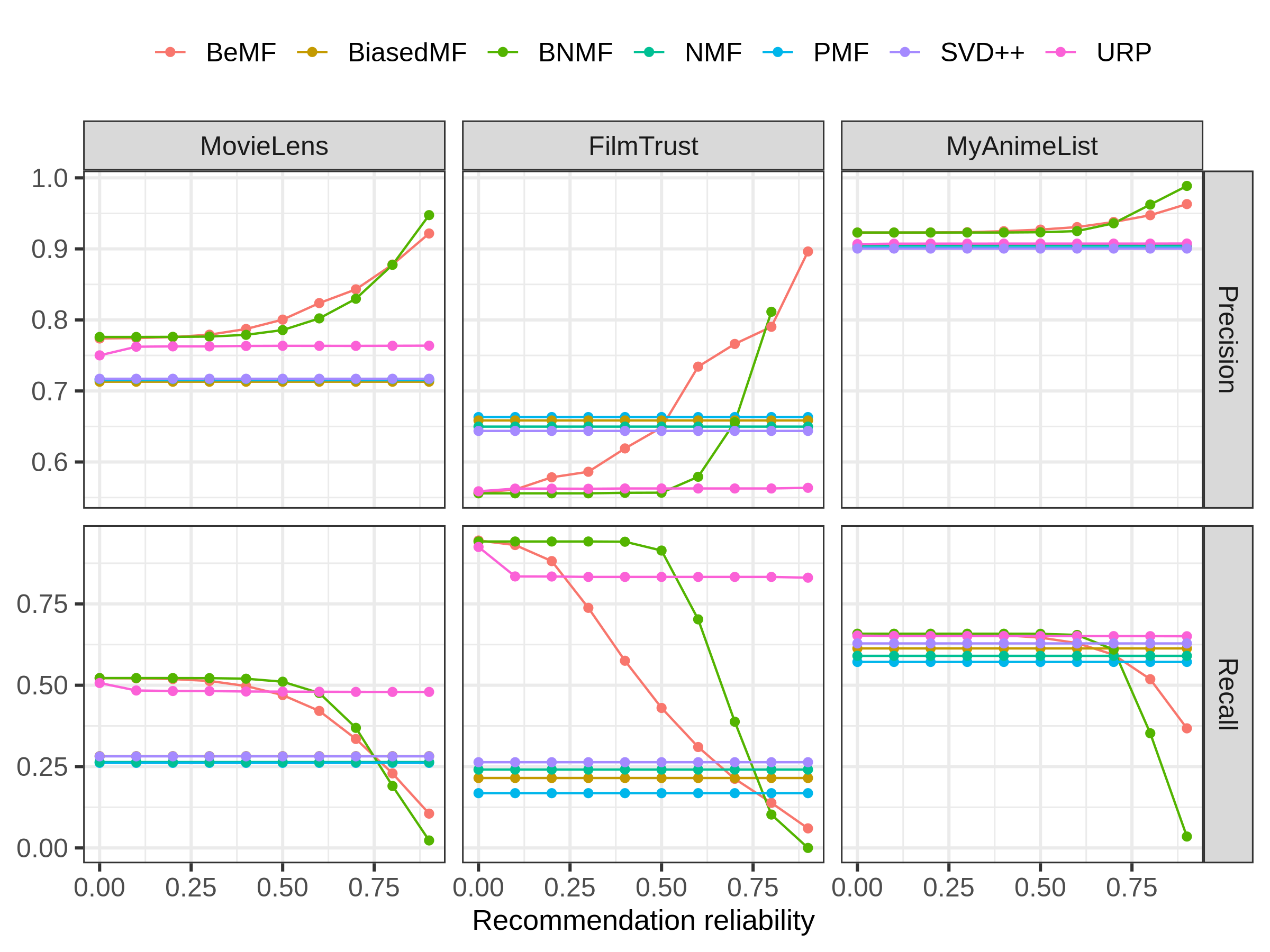}
    \caption{Quality of the recommendations measured by precision and recall. Recommendations with a lower probability than those denoted on the $x$-axis are filtered out.}
    \label{fig:precision-vs-recall}
\end{figure}

Recommendation lists, including the top 10 test items, were built in two ways. If the recommendation method provides the reliability of the recommendations (refer to \cref{tab:baselines-outputs}), we have selected the top 10 test items with the highest probability of being liked by the user (reliability $\rho_{u,i} \geq \vartheta$). On the other hand, if the recommendation method does not provide the reliability of the recommendations, we have selected the top 10 test items with the highest prediction ($\hat{R}_{u,i}$), excluding those who have a prediction lower than $\theta$. Note that $\vartheta$ and $\theta$ have different roles: the former is a threshold in the desired reliability, while the latter is a threshold in the predicted score. The results of this comparison are shown in \cref{fig:precision-vs-recall}.

This plot shows that when recommendations with a lower probability than those denoted on the $x$-axis are filtered out, the precision value increases and the recall value declines. \ac{BeMF} is the recommendation method that provides the best precision when the item’s minimum probability of being liked is between $0.5$ and $0.75$. \ac{BNMF} achieves the best precision with probabilities higher than $0.75$. However, the recall value is so low in these cases that the recommendation ability of the method is seriously impaired. Therefore, \ac{BeMF} provides a better balance between the quality of the recommendations (precision) and the capability of the recommendation items (recall).

\section{Conclusions}\label{sec:conclusions}

In this paper, we have presented the \ac{BeMF} model, which is an \ac{MF}-based \ac{CF} algorithm that returns not only predictions for items that are not rated by users but also the reliability of these predictions. This outcome is achieved by addressing the recommendation process as a classification problem rather than a regression problem. The pair $\langle \textrm{prediction}, \textrm{reliability}\rangle$ that is returned by the  \ac{BeMF} model allows us to calibrate its output to obtain the proper balance between the quality of the predictions and the quantity of the predictions (i.e., to decrease the prediction error by reducing the model’s coverage). Likewise, the  \ac{BeMF} model allows us to estimate the reliability of a recommendation by obtaining the probability that a user has interest in an item. This approach is innovative approach that improves accuracy by selecting recommendations with the highest reliability values and providing explicit reliability values to the users.

The experimental results carried out on the MovieLens, FilmTrust and MyAnimeList datasets show the distinct superiority of the \ac{BeMF} model against not only other \ac{MF} models but also alternative approaches such as \ac{KNN} when filtering out unreliable predictions and recommendations. \ac{BeMF} has also proven to achieve a better balance between the quality of the recommendations (precision) and the capability of recommending items (recall). The authors of this work are committed to reproducible science, so the source code of all the experiments conducted in this article is publicly accessible and was tested against benchmark datasets that are included in the open source project \ac{CF4J}.

One of the most important reasons for this outperformance is that reliability is a native concept in \ac{BeMF} and does not depend on external methods or extended architectures as existing solutions. Using the Bernoulli distribution, \ac{BeMF} can model the rating action as a stochastic process, in which the decision of assigning a particular score is not deterministic but depends on some psychologically and environmentally imponderable factors of the users (e.g., its frame of mind) and the items (e.g., hype around a long-awaited movie). In this way, reliability is intrinsically linked to \ac{BeMF}, and in some sense, is the main focus of the algorithm: prediction is just a byproduct that is obtained by choosing the most reliable score. This focus is in sharp contrast with reliability measures that can be artificially added to other methods, which are based on either a secondary \ac{MF} to predict the expected error~\cite{zhu2018assigning} or a similarity measure that is computed on the selected neighbors~\cite{bobadilla2010new}.

On the other hand, the \ac{URP} returns a native reliability measure, which is provided via a probabilistic model of the behaviors of the users (essentially a multinomial distribution). However, in this case, the proposed distribution imposes strong topological restrictions on the obtained reliability measures. If the \ac{URP} returns that the most likely score for an item from a given user is $s \in \mathcal{S}$, then the shape of the multinomial distribution forces that $s-1$ and $s+1$ are also very likely ratings. In particular, the obtained distributions are unimodal, so reliability is essentially a measure of closeness to the mode. Conversely, BeMF supports any discrete distribution, including multimodal distributions, as output. This finding implies that \ac{BeMF}'s reliability is much richer, and it may reach a valley between two modes, e.g., if an item is very controversial and is usually rated very well or very bad.

In future work, we propose conducting a more in-depth analysis of the impact of the proposed model beyond the quality of the predictions and recommendations. We propose studying the quality of \ac{BeMF} beyond accuracy quality measures, such as novelty, diversity or discovery. Similarly, we propose to study the stability of the model against shilling attacks that are performed to nuke or promote particular items.

Another interesting research area is the use of other probability distributions as underlying assumptions for the \ac{MF} method. In this paper, we modeled the ratings as a $D$-dimensional vector of independent Bernoulli random variables, which enables efficient parallel training for each possible rating and reduces the number of hyperparameters of the model. Nevertheless, some dependency among random scores can be imposed. This dependency, such as modeling votes as a normal distribution in \ac{PMF}~\cite{mnih2008probabilistic} or as a binomial distribution in \ac{BNMF}~\cite{Hernando2016Apr}, has been previously reported in the literature. A prospective work would be to explore other types of distributions that allow us to improve the performance in specific tasks. For instance, modeling the ratings as an exponential distribution allows the model to capture and exploit rare events, such as recommending underrated movies to specific users or providing accurate predictions to cold users.

On the other hand, this new model can be extended. Therefore, we propose incorporating social and content information to improve the quality of the predictions. We also propose adapting \ac{BeMF} to make recommendations to user groups. An interesting research area would be to add time information to the model to reflect the changes in the opinions of the users over time.

\section*{Acknowledgments}

This work has been co-funded by the \textit{Ministerio de Ciencia e Innovación} of Spain and European Regional Development Fund (FEDER) under grants PID2019-106493RB-I00 (DL-CEMG) and TIN2017-85727-C4-3-P (DeepBio). The authors thank Dolores Abernathy for very useful conversations regarding the application of logistic functions to the model.


\bibliographystyle{abbrvnat}
\bibliography{sample.bib}

\end{document}